\documentclass[final,5p,times,twocolumn]{elsarticle}

\usepackage{graphicx}
\usepackage{amssymb}
\usepackage{float}
\usepackage{placeins}
\usepackage{amsmath}
\usepackage[table]{xcolor}
\usepackage{array}
\usepackage{multirow}
\usepackage{booktabs}
\usepackage[group-separator={,}]{siunitx}

\newcommand\MyBox[2]{
  \fbox{\lower0.75cm
    \vbox to 1.7cm{\vfil
      \hbox to 1.7cm{\hfil\parbox{1.4cm}{#1\\#2}\hfil}
      \vfil}%
  }%
}

\usepackage{amsthm}

\usepackage{lineno}
\journal{Expert Systems with Applications}

\begin{document}

\begin{frontmatter}


\title{Minimizing the Societal Cost of Credit Card Fraud with Limited and Imbalanced Data}



\author{Samuel Showalter, Zhixin Wu}
\address{Department of Mathematics, DePauw University, Greencastle, IN 46135, USA}

%%%%%%%%%%%%%%%%%%%%%%%%%%%%%%%%%%%%%%%%%%%%%%%%%%%%%%%%%%%%%%%%%%%%%%%%%%%%%%%%%%%%%%%%%%%
% Abstract Section
%%%%%%%%%%%%%%%%%%%%%%%%%%%%%%%%%%%%%%%%%%%%%%%%%%%%%%%%%%%%%%%%%%%%%%%%%%%%%%%%%%%%%%%%%%%
%Abstract Section
\begin{abstract}
Machine learning has automated much of financial fraud detection, notifying firms of -- or even blocking -- questionable transactions instantly. However, data imbalance starves traditionally trained models of the content necessary to detect fraud. This study examines three separate factors of credit card fraud detection via machine learning. First, it assesses the potential for different sampling methods -- undersampling and Synthetic Minority Oversampling Technique (SMOTE) -- to improve algorithm performance in data-starved environments. Additionally, five industry-practical machine learning algorithms are evaluated on total fraud cost savings in addition to traditional statistical metrics. Finally, an ensemble of individual models is trained with a genetic algorithm to attempt to generate higher cost efficiency than its components. Monte Carlo performance distributions discerned random undersampling outperformed SMOTE in lowering fraud costs, and that an ensemble was unable to outperform its individual parts. Most notably, the $F_1$ Score, a traditional metric often used to measure performance with imbalanced data, was uncorrelated with derived cost efficiency. Assuming a realistic cost structure can be derived, cost-based metrics provide an essential supplement to objective statistical evaluation.
\end{abstract}


\end{frontmatter}

%\linenumbers

%%%%%%%%%%%%%%%%%%%%%%%%%%%%%%%%%%%%%%%%%%%%%%%%%%%%%%%%%%%%%%%%%%%%%%%%%%%%%%%%%%%%%%%%%%%
% Introduction section
%%%%%%%%%%%%%%%%%%%%%%%%%%%%%%%%%%%%%%%%%%%%%%%%%%%%%%%%%%%%%%%%%%%%%%%%%%%%%%%%%%%%%%%%%%%

\section{Introduction}
\label{S:1}

%General information about fraud detection and the problems inherent in classifying
Since the birth of e-commerce in the early 1990s, Internet purchasing and credit card fraud have proliferated across the globe. Moreover, the host of challenges inherent in detecting credit card fraud has made the field one of the most explored \cite{Phua2010}. Outlined in \cite{DalPozzolo2017}, individual purchasing behavior evolves over time, as do the methods with which people commit fraud. Credit card transactions are both high in volume and imbalanced \cite{DalPozzolo2014New}, with few genuinely fraudulent transactions. High volume also delays transaction feedback, prolonging the time needed to adequately verify and categorize questionable activity \cite{DalPozzolo2017}.

%History of fraud detection and the costs associated with fraud
Originally, manual oversight and client feedback were the only defenses against fraud attacks. Today, however, a multitude of rule-based and machine learning algorithms exist that constantly scan for contentious credit card purchases \cite{Phua2010,Kuo2004,Bolton2002,Aleskerov1997}. Despite these preventative measures, a 2016 LexisNexis survey \cite{LexisNexis2016} of hundreds of corporations approximated fraud cost to be billions annually. On average, businesses estimated their annual cost of fraud to be $1.47\%$ of total revenues.  Furthermore, fraud occurrence is substantially higher for online retailers \cite{LexisNexis2016}, and increased domination of mobile and e-pay options was the single most cited concern companies had about preventing fraud. Over 48\% of online fraudulent transactions were made with a credit card in 2016, far higher than debit card transactions \cite{LexisNexis2016} and other methods.

%Brief overview of the research and the types of credit card fraud
Dozens endeavor to optimize the detection of fraud \cite{Ngai2011}. While credit card fraud is particularly well researched, it is only one of many documented types spanning multiple industries. Even so, its impact is disproportionately large. Credit card fraud can either take the form of \textit{application} or \textit{behavioral} fraud. Application fraud is the act of procuring a credit card using falsified or stolen information, and can be thought of as a type of identity theft. Behavioral fraud, a much more common occurrence, is the act of purchasing goods with stolen credit card information. While this can occur at Point of Sale (POS) locations, it is more commonly seen in Internet transactions \cite{LexisNexis2016}. Many sources of information are leveraged to predict behavioral fraud, but profiling the purchasing activity of individuals provides a functional benchmark from which future transactions can be compared \cite{Delamaire2009}. Temporal and location-based information is also useful in this regard. 

%Talk about downsides and limitations of fraud detection, moving in to scarcity of data.
However, if purchasing behavior is sporadic or scant, profiling is less useful. Consistently insightful transaction traits are difficult to find, as is procuring data to test different Fraud Detection Systems (FDS). Credit card transactions are a form of Personally Identifiable Information (\textit{PII}), dictating many features must be hashed or encoded. Additionally, banks and retailers alike are extremely secretive about fraud \cite{Phua2010} due to its sensitive and unsavory nature. The lack of fraud transparency in the public and private sectors is a consistent point of criticism.

% Talk about the machine learning models commonly implemented
In spite of data scarcity, countless Fraud Detection Systems have been created in the past decade. Generally, these are categorized as either supervised or unsupervised \cite{Kuo2004}. Supervised machine learning models face a number of challenges, namely an inability to identify new types of fraud as well as a perpetual delay of data in what is known as $verification \ latency$ \cite{DalPozzolo2017}. Questionable activity is often manually checked, and the resulting delay is often ignored by fraud detection studies \cite{DalPozzolo2017}. Conversely, unsupervised algorithms tend to cluster transactions based on similarity. This enables a FDS to potentially categorize new types of fraud, though human intervention is needed to characterize each cluster \cite{Bolton2002}. In general, fraud detection has not yet evolved enough to detect fraud independently. Rather, anomaly detector may be a more adequate characterization of these techniques.

%Discuss the different sampling methods used in this space
Further complicating the matter, fraudulent activity exists as a minute percentage of total credit card transactions, at times less than 0.2\% of available data. The predictive power of fraud systems is heavily impacted by the composition of its training data. To combat classification bias, alternative methods re-balance sample composition in favor of minority classification. One of the most promising of these methods is undersampling, which randomly chooses a subsample of the majority class, placing emphasis on the minority class. Conversely, oversampling seeks to accomplish the same goal by sampling with replacement from a minority class up to a certain threshold \cite{Chawla2002}.

%SMOTE information
Likewise, a novel sampling technique developed by \cite{Chawla2002} incorporates facets of both undersampling and oversampling. Synthetic Minority Oversampling Technique (SMOTE) can be described as randomly undersampling the majority class as well as generating synthetic minority records. While SMOTE's technique is novel, Blagus et. al. \cite{Blagus2013} finds it does not outperform undersampling when applied to gene expression data. According to their empirical results, synthetic records beyond a given threshold confound the classification boundary between the majority and minority classes.

%Mention the different ways of measuring performance and the debates therein (brief)
Lastly, an ongoing debate examines the ideal method of evaluating fraud detection algorithms. High data imbalance causes accuracy and other traditional metrics to be misleading, as ignoring the minority class entirely could yield an accuracy well over 99\%. Many have posed alternative metrics \cite{van2015apate}, including cost-based indicators and ranking systems.

%%%%%%%%%%%%%%%%%%%%%%%%%%%%%%%%%%%%%%%%%%%%%%%%%%%%%%%%%%%%%%%%%%%%%%%%%%%%%%%%%%%%%%%%%%%
% Contribution section
%%%%%%%%%%%%%%%%%%%%%%%%%%%%%%%%%%%%%%%%%%%%%%%%%%%%%%%%%%%%%%%%%%%%%%%%%%%%%%%%%%%%%%%%%%%
\section{Contribution}
\label{S:2}

%Overview of objectives as well as sampling methods overview
This paper has three objectives, all of which pertain to optimizing fraud detection in data-starved environments. The dataset implemented to test these theories is public, anonymized, and has been transformed by Principal Component Analysis \cite{DalPozzolo2015}. Therefore, longitudinal data is absent and temporal and location-based profiling techniques are not possible. Accordingly, this study compares the efficacy of undersampling and SMOTE in training machine learning models. Since standard sampling is ineffective with imbalanced data \cite{Chawla2010}, it is crucial to understand the predictive potential of alternative methods. 

%Comparison of traditional performance metrics and cost-based evaluation
Additionally, by tuning and testing a variety of commonly used machine learning classifiers, this paper explores the correlation between traditional performance metrics (Positive Predictive Value ($PPV$), True Positive Rate ($TPR$), $F_1$ Score) and cost efficiency. Though previously examined by \cite{Chan1999} \cite{Elkan2001} \cite{Phua2004}, economic cost evaluation is relatively unexplored despite promising initial findings \cite{Bahnsen2013}. In the same vein, examining fraud cost in situations with little data is particularly  crucial as new forms of fraud may not be documented. Our findings serve as a contemporary exploration of previous research, but with a greater focus on societal cost ease of industrial adoption.

%Genetic algorithm information
Finally, genetic algorithms train an ensemble classifier comprised of independent algorithms. Preeminent models from previous experiments are combined and assigned weights. Iteratively, a genetic algorithm optimizes these weights within specified threshold bounds. Ultimately, this experiment discerns the feasibility of improving fraud detection through a multi-model voting system. Such a practice could prove useful when little information can be gleaned from transaction history alone.

%%%%%%%%%%%%%%%%%%%%%%%%%%%%%%%%%%%%%%%%%%%%%%%%%%%%%%%%%%%%%%%%%%%%%%%%%%%%%%%%%%%%%%%%%%%
% Dataset information section
%%%%%%%%%%%%%%%%%%%%%%%%%%%%%%%%%%%%%%%%%%%%%%%%%%%%%%%%%%%%%%%%%%%%%%%%%%%%%%%%%%%%%%%%%%%
\section{Procuring Publicly Available Credit Card Data}
\label{S:3}

%General overview of the data used
The paper utilizes a dataset comprised of 284,807 credit card transactions occurring over two days. Published by unnamed European bank in September 2013 \cite{DalPozzolo2015}, only 0.172\% (492 records) of this data was fraudulent. As one of the few publicly released fraud datasets, the features of these transactions were anonymized (as $V_1, V_2, ... V_n$) and transformed using Principal Component Analysis (PCA) \cite{jolliffe1986principal} before being posted publicly. PCA de-dimensionalizes data such that the only the most distinguishable traits remain, implementing orthogonal transformation to convert a set of observations with correlated variables into a linearly uncorrelated set. Only the $amount$ and $time$ variables were not transformed.

%Implications of the data
As a result, connecting transactions to a specific user is impossible. Even if data was not anonymized, transaction history over a two day window is likely unhelpful with profile based classification. Therefore, the central focus of this paper is to identify optimal sampling methods and model training practices in the absence of situational context or historical data.

%%%%%%%%%%%%%%%%%%%%%%%%%%%%%%%%%%%%%%%%%%%%%%%%%%%%%%%%%%%%%%%%%%%%%%%%%%%%%%%%%%%%%%%%%%%
% Sampling Methods section
%%%%%%%%%%%%%%%%%%%%%%%%%%%%%%%%%%%%%%%%%%%%%%%%%%%%%%%%%%%%%%%%%%%%%%%%%%%%%%%%%%%%%%%%%%%
\section{Sampling Methods}
\label{S:4}

% Simple random sample subsection -- used as the baseline metric
\subsection{Simple Random Sampling}

Traditionally, machine learning models are trained on a simple random sample of population data. Aside from preprocessing, often all that is necessary to prepare data with popular machine learning software \cite{pedregosa2011scikit} is a test ratio, or the percentage of data to be used for testing classifier performance. However, it is commonly found that \cite{Chawla2002,DalPozzolo2015,Chawla2010} simple random sampling is unhelpful with problems of severe data imbalance. In turn, alternative methods that artificially rebalance data are almost exclusively used in industry and academia.

% Undersampling subsection -- used and tested in the paper
\subsection{Undersampling}

Liu et. al. \cite{liu2009}  describes the process of undersampling as using a ``subset of the majority class to train [a] classifier". In this manner, classifiers are more directly influenced by the characteristics of the minority class. In practice, \cite{drummond2003} confirms that undersampling outperforms random oversampling, a method that, instead of collecting a subset of the majority, generates a sample of minority records by selecting with replacement. Applications of oversampling tend to suffer from an ill-defined minority classification boundary. With-replacement samples of the minority class, regardless of size, often cannot generalize trends beyond those present in training data \cite{Chawla2010}. However, undersampling suffers from its own pitfalls. If the chosen subset of majority class data is not prescriptive enough to identify the majority population, the classifier will be unable to generalize majority class identification. Even so, \cite{liu2009} finds that implementing a bagging ensemble trained with different majority subsets assuages the risks of information omission in training. This innovation, coupled with ancillary techniques, have made undersampling a staple of imbalanced data analysis.

% SMOTE subsection -- used and tested in the paper
\subsection{SMOTE: Synthetic Minority Oversampling Technique}

A combination of undersampling and oversampling is known as Synthetic Minority Oversampling. SMOTE endeavors to compensate for undersampling's weaknesses by allowing for larger majority sample sizes while maintaining a specified ratio of minority records. It is also thought to assist in generalizing a minority feature space with few records \cite{Chawla2002}.  To do so, a minority record is randomly chosen, as are its most similar $k$ ``neighbors" using euclidean distance or an equivalent similarity measure. A second minority record is then randomly chosen from the  $k$-neighbor subsample. Features $X_{1 - n}$ for the synthetic record are then derived as a linear combination of the original record $\alpha_i$ and its difference with selected neighbor $\alpha_k$, multiplied by a random  scalar $c$ chosen from a uniform distribution.
\begin{align*}\label{eq:SMOTE}
{X_i = \alpha_i + c(\alpha_i - \alpha_k)} \ \ \ \ \  \ \ \ \ \ \ i \  =  \ 1,2, ...  \ n \\
{c \ \sim \ U(0,1)}
\end{align*}

Chawla (2002) \cite{Chawla2002} provides evidence that supports the validity of SMOTE's aforementioned merits. Asserting that synthetically generating records prevents ``overfitting on the multiple copies of minority class examples", Chawla (2002) maintains SMOTE causes the ``decision region of the minority class to become more general." At the same time, increasing the number of minority records through synthetic creation allows for a greater number of majority records to be included in the training data while maintaining the desired minority-majority ratio. Ramezankhani et. al. (2016) \cite{Ramezankhani2016} also found boosted performance with SMOTE in lipid and glucose experimentation.
%%%%%%%%%%%%%%%%%%%%%%%%%%%%%%%%%%%%%%%%%%%%%%%%%%%%%%%%%%%%%%%%%%%%%%%%%%%%%%%%%%%%%%%%%%%
% Determining model performance section
%%%%%%%%%%%%%%%%%%%%%%%%%%%%%%%%%%%%%%%%%%%%%%%%%%%%%%%%%%%%%%%%%%%%%%%%%%%%%%%%%%%%%%%%%%%
\section{Calculating Model Performance with Data Imbalance }
\label{S:5}

As traditional sampling techniques falter with imbalanced optimization problems \cite{Blagus2013}, so do traditional evaluation strategies. Other than accuracy measures, a common practice in binary classification problems is to define performance in terms of $true \ positive$, $false \ positive$, $true \ negative$, and $false \ negative$ rates. As seen in the $confusion \ matrix$ below, true positives and true negatives are correctly classified fraudulent and non-fraudulent records, respectively. By contrast, a false negative occurs when fraud goes undetected and a false positive characterizes a false alarm, when a transaction is flagged but not fraudulent. Segmenting classifier performance into these four groups affords researchers the freedom to weight different outcomes independently \cite{DalPozzolo2017}.

% %Confusion Matrix
\begin{table}[h]
\begin{tabular}{c >{}r @{\hspace{0.7em}}c @{\hspace{0.4em}}c @{\hspace{0.7em}}l}
  \noindent
  \renewcommand\arraystretch{1.5}
  \setlength\tabcolsep{0pt}
  \multirow{11}{*}{\parbox{0.7cm}{\bfseries\raggedright Prediction}} & 
    & \multicolumn{2}{c}{\bfseries Actual Transaction} & \\
  & Fraud & \MyBox{True}{Positive (TP)} & \MyBox{False}{Positive (FP)} \\[2.4em]
  & {Not Fraud} & \MyBox{False}{Negative (FN)} & \MyBox{True}{Negative (TN)} \\
 && Fraud & Not Fraud &
\end{tabular}
\caption{Confusion matrix for credit card fraud detection}
\end{table}

Subsequently, many assessment metrics are derived from table 1. Listed in equations \ref{eq:TPRFPR}, \ref{eq:TNRFNR}, \ref{eq:PPVFDR}, and \ref{eq:NPVFOR} are the eight basic indicators derived from confusion matrices: true and false positive and negative rates ($TPR$, $FPR$, $TNR$, $FNR$), positive and negative predictive value ($PPV$, $NPV$), false discovery rate ($FDR$), and false omission rate ($FOR$). Visually, all equations in the left column focus on successful classification, while the right column focuses on failure rates.

\begin{equation}\label{eq:TPRFPR}
{TPR = \frac{TP}{TP + FN} \ \ \ \ \ \ \ \ \ \ FPR = \frac{FP}{FP + TN} }
\end{equation}

\begin{equation}\label{eq:TNRFNR}
{TNR = \frac{TN}{TN + FP} \ \ \ \ \ \ \ \ \ \ FNR = \frac{FN}{FN + TP} }
\end{equation}

\begin{equation}\label{eq:PPVFDR}
{PPV = \frac{TP}{TP + FP} \ \ \ \ \ \ \ \ \ \ FDR = \frac{FP}{TP + FP} }
\end{equation}

\begin{equation}\label{eq:NPVFOR}
{NPV = \frac{TN}{TN + FN} \ \ \ \ \ \ \ \ \ \ FOR = \frac{FN}{TN + FN} }
\end{equation}

\bigbreak
%TPR and TNR versus PPV and NPV
Conceptually, TPR and TNR determine how many fraud records were appropriately classified. Both of these metrics are direct evaluations of model performance by class, and ask the question: \textit{how many fraud records were classified correctly?} Conversely, PPV and NPV can be described as a $prediction \ centric$ evaluation of the confusion matrix, conditionally examining results given a specific prediction. That is, \textit{for all records predicted to be fraudulent, how many actually were?}  

%Discussion of TPR, FNR, and AUROC
One way these indicators can discern classifier performance is with the Area Under the Reciever Operating Characteristic (AUROC) \cite{bewick2004statistics}. Implemented to increase emphasis on detecting the minority class, AUROC maps an algorithm's  $TPR$ and $FPR$ classification ability at different thresholds. However, as noted in \cite{Bahnsen2016}, some confusion matrix measures are biased for imbalanced datasets due to a disproportionally large number of true negative records. False Positive Rates (\textit{FPR}) are held artificially low by this imbalance, over-estimating fraud detection efficacy.

%Explanation of 
To garner a more realistic picture of performance in an imbalanced setting, researchers derived the Area Under the Precision Recall (AUPRC) curve. Subtly distinct from AUROC, AUPRC maps $TPR$ and $PPV$, also known as recall and precision. AUPRC is thought to assuage issues of \textit{FPR} bias by considering the \textit{success rate} of minority classification rather than its failure rate, correcting biases caused by the majority class. Indeed, Davis \& Goadrich (2006) \cite{davis2006relationship} assert that a "curve dominates in ROC space if and only if it dominates in PR space."  That is, AUPRC measurements embody a more robust method of testing for algorithmic optimality. 

%Discussion of cost-based metrics
Nevertheless, some still take issue with AUPRC as the definitive performance indicator as the ultimate pursuit of fraud detection is to minimize the cost to society \cite{Elkan2001}. As noted by \cite{LexisNexis2016}, this cost takes different forms. Implications beyond the de-facto cost of the fraudulent transaction include the time-loss on behalf of the bank or retailer in identifying the issue, loss of trust (and potentially business) of clients and customers, and loss of productivity on behalf of the victim who may need to apply for a new card. The fraud multiplier of $\$2.40$ \cite{LexisNexis2016} only covers cost incurred by the company, excluding cost to the individual victim or tertiary party involved. Thus, \cite{Chan1999},  \cite{Bahnsen2013}, and others propose a derived cost function.

%Cost matrix
\begin{table}[h]
\begin{tabular}{c >{}r @{\hspace{0.7em}}c @{\hspace{0.4em}}c @{\hspace{0.7em}}l}
  \noindent
  \renewcommand\arraystretch{1.5}
  \setlength\tabcolsep{0pt}
  \multirow{11}{*}{\parbox{0.7cm}{\bfseries\raggedright Prediction Cost}} & 
    & \multicolumn{2}{c}{\bfseries Actual Transaction Cost} & \\
  & Fraud & \MyBox{$-T_c + C_{f}$}{} & \MyBox{\Large{ \ \ \ $C_e$}}{} \\[2.4em]
  & {Not Fraud} & \MyBox{$F_m(T_c) + C_{l}$}{} & \MyBox{ \ \ \ \ \ \  \Large{0}}{} \\
 && Fraud & Not Fraud &
\end{tabular}
\caption{Cost matrix for evaluating credit card fraud detection algorithms.}
\end{table}

%Explanation of the cost matrix
Cost matrices allow the user to dynamically weight the importance of different outcomes. Table 2 represents a unique cost matrix similar to those proposed by \cite{Chan1999} and \cite{Bahnsen2013}. Identifying and preventing a fraudulent transaction equates to saving the cost of the transaction $-T_c$ as well as the time cost $C_{f}$ of resolving the event. Similarly, false alarms ($FP$) cost the amount of time needed to correct the error, $C_e$. Most importantly, failing to detect fraud increases the cost of the transaction by the fraud multiplier $F_m$, plus the cost of addressing the loss, $C_l$.

%Discussion of the debate around the cost matrix and other metrics
The sensibility of cost-based evaluation has been questioned since its inception despite its practicality \cite{Chan1999}. Simply put by \cite{DalPozzolo2014}, ``the cost of a fraud is not easy to define." Indeed, contemporary Fraud Detection Systems have multiple layers. Rule-based credential authorization offers a first layer of defense, followed by algorithmic classification and terminating at human oversight. Derivation of company-specific cost models is hampered by this complexity. Additionally, \cite{DalPozzolo2017} asserts that cost-based metrics ignore the disruptive nature of fraud alerts and company costs. However, such an argument also applies to many traditional metrics, including AUPRC. Debate continues and traditional metrics are still implemented as performance indicators \cite{Phua2010}.

%%%%%%%%%%%%%%%%%%%%%%%%%%%%%%%%%%%%%%%%%%%%%%%%%%%%%%%%%%%%%%%%%%%%%%%%%%%%%%%%%%%%%%%%%%%
% Related works (lit. review of sorts) section
%%%%%%%%%%%%%%%%%%%%%%%%%%%%%%%%%%%%%%%%%%%%%%%%%%%%%%%%%%%%%%%%%%%%%%%%%%%%%%%%%%%%%%%%%%%
\section{Evolution of Fraud Detection Systems}
\label{S:6}
%General overview of related works on fraud detection
In the past two decades, roughly a dozen articles summarize hundreds of experiments conducted on different types of fraud. Many of these dedicate large portions of their articles to credit card fraud \cite{Delamaire2009,Kuo2004, Ngai2011, Phua2010}. Of note, \cite{Ngai2011} identify outlier detection via supervised learning to be the most common, though it is implemented in many different forms.

%Exploration of neural networks
In particular, neural networks are included in a plurality of Fraud Detection Systems. In fact, neural network applications have been utilized since the proliferation of e-commerce. Compensating for data imbalance with Fisher's linear discriminant analysis, \cite{dorronsoro1997neural} created a neural network classifier that stood as a cornerstone achievement alongside CARDWATCH \cite{Aleskerov1997}. Today, some networks make use of unsupervised clustering models to identify unprecedented fraud methods. Even with this advantage, $verification \ latency$ stands as a major drawback for all supervised models, particularly neural networks that require significant time to train.

%Mention novel new strategies (short)
The additional challenge of \textit{concept drift} and\textit{ verification latency}, along with the increasing cost of fraud, have led to the creation of innovative new techniques spanning multiple disciplines. Graph-based anomaly detection and inductive logic are only a few new implementations \cite{Phua2010}. Multi-faceted ranking, searching, and scoring ensemble methods have also spread in the past decade \cite{DalPozzolo2014}, often coupled with cost-based evaluation and Genetic Algorithms \cite{gadi2008credit}.

%Discuss the inability of models to actually be implemented in industry (resource req., not cost related...)
%Academics also tend to overcomplicate the models
Unfortunately, systems derived in academia do not consistently transfer to a corporate setting.  While contemporary models continue to increase in complexity, Phua et. al. (2010) \cite{Phua2010} laments that many fraud-plagued industries are "dependent on the practical issues of operational requirements, resource constraints, and management commitment towards reduction of fraud." Such constraints are rarely, if ever, imposed on models in a research setting. Despite the fact that experiments on computationally efficient algorithms \cite{Bhattacharyya2011} are continually conducted and published, their industrial practicality is seldom emphasized. In turn, the following experiment will only make use of relatively lightweight, scalable classification methods.

%%%%%%%%%%%%%%%%%%%%%%%%%%%%%%%%%%%%%%%%%%%%%%%%%%%%%%%%%%%%%%%%%%%%%%%%%%%%%%%%%%%%%%%%%%%
%
% END OF THE SECTIONS WHERE I NEED TO SITE THINGS A LOT AND THINK REAL HARD, YAY!
% Also, start of Methods section
%%%%%%%%%%%%%%%%%%%%%%%%%%%%%%%%%%%%%%%%%%%%%%%%%%%%%%%%%%%%%%%%%%%%%%%%%%%%%%%%%%%%%%%%%%%

\section{Research Methods}
\label{S:7}

% Discussion of the methods used to automate the testing (my software!)
\subsection{Automating Fraud Detection Experiments}

To examine a variety of samples and models, as well as train an ensemble with genetic algorithm (GA) optimization, a modular testing platform was created. Encapsulated in $Test$ objects, a specified sample and model are run repeatedly via Monte Carlo simulations, as shown in Figure 1. Results are stored after each testing cycle in a $Logger$ object as a result log before the sample is re-created and the process repeats. If an ensemble is specified, performance will be stored in the same location. At the end of the execution, a master log of all execution parameters, average results, and meta-data is sent to the $Logger$.

%Diagram of fraud detection model engine
\begin{figure}[h]
\label{fig:platform}
\centering\includegraphics[width=1\linewidth]{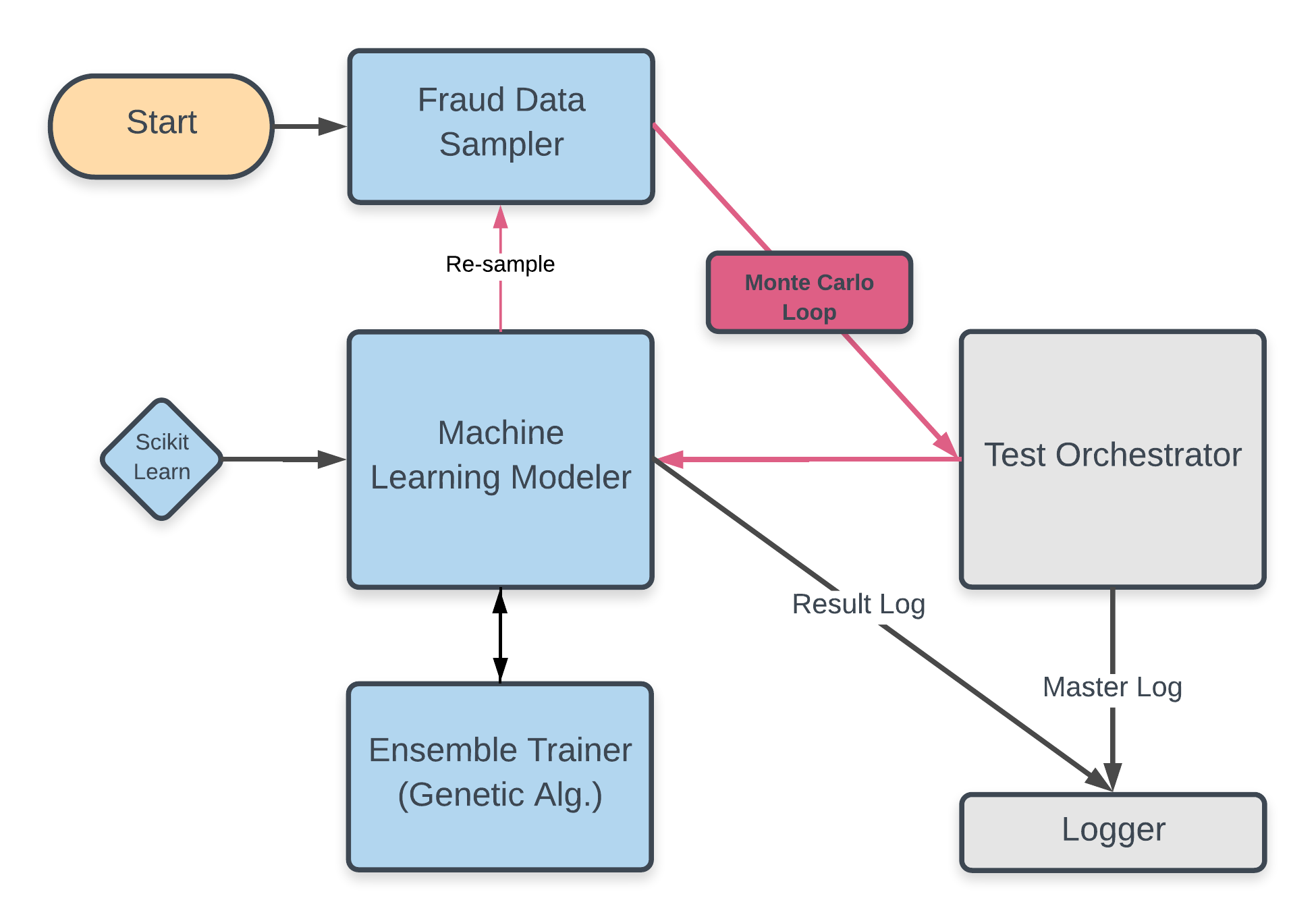}
\caption{Scalable automation platform for testing fraud detection algorithms}
\end{figure}

\textbf{Sampling:} All executions begin with a command sent to the sampling engine that randomly selects 20\% of the entire dataset to be used for sample creation. The remaining records are set aside for testing. Fed approximately 100 and 59,900 fraud and non-fraud records, respectively, the sample engine then launches an orchestration package dependent on sample method, sample size, and fraud ratio parameters. Undersamping samples ignore the desired sample size, as its maximum size is constrained by the desired ratio of fraud records. Conversely, SMOTE samples includes all fraudulent records, and then synthetically adds to the minority subsample until the specified fraud threshold is reached. Any records not used in the final sample are added to the testing dataset. Due to PCA transformations of the source data, no preprocessing is considered aside from sample generation. 

\bigbreak
\textbf{Modeling:} Provided a training sample, the machine learning modeler enters a Monte Carlo loop with a specified number of iterations.  Implemented with Sci-kit Learn, the modeler tests Logistic Regression, Linear Support Vector, Random Forest, K-Nearest Neighbors, and Gaussian Naive Baye's classifiers. KNN serves as the control due to its notoriously poor performance with imbalanced datasets \cite{Bolton2001}. Each model is trained and tested on the same sample, with performance metrics logged ad-hoc. After the execution terminates, the modeler re-instantiates the sample object, which re-partitions the entire dataset and creates a new sample. Once the Monte Carlo loop completes, performance records for each model are averaged and stored in the master log.

\bigbreak
%Overview of why ensembles are created
\textbf{Ensemble:} If little information can be gleaned from a dataset, then coordinating multiple predictions may yield a better result. Just as a governing body operates on popular vote, so too does the fraud detection ensemble. By assigning weights to different fraud predictors and then optimizing them with genetic algorithms, ensemble performance may exceed that of its components.

%How do you test these theories about ensembles -- what changes for regular execution
To test this theory, the modeler engine is provided an ensemble evolution object. If desired, this object is instantiated and called as a part of the Monte-Carlo cycle. Feedback during evolution is provided by partitioning the original sample into a training and testing subsets. All individual models are trained with the training subsample, and their individual predictions are recorded.

%Now the cool part about evolution
These individual predictions are then assigned random weights, and genetic evolution commences. A random population of weights is generated, and those with the highest fitness (lowest cost) receive a increased probability of being chosen to populate the following generation. New generations of weights inherit traits from their parents through bit string crossover. Fitness is then calculated for the new offspring, and the process continues until the execution times out. After evolution terminates, individual models generate predictions for the testing sample. Optimized weights are applied to these transactions, giving a weighted average probability of fraud. 

%Include equations for how genetic algorithm works here!!!

\subsection{Cost-Based and Statistical Evaluation Practices }

During executions, $TP$, $FN$, $TN$, and $FP$ rates are stored for each model, from which many additional metrics ($TPR$, $FNR$, $PPV$, $NPV$) are derived. Furthermore, general $Precision$ and $Recall$ metrics shown in equation \ref{eq:PrecRecall} provide holistic measure of proficiency by class ($Recall$) as well as by prediction ($Precision$). These are composite competency metrics, not to be confused with $PPV$ and $TPR$, sometimes referred to as precision and recall. Contextually, $Precision$ here is defined as the mean prediction precision: what percentage of each prediction type were correct? Similarly, $Recall$ discerns the mean percentage of each class that were categorized correctly.

\begin{equation}\label{eq:PrecRecall}
{Precision = \frac{PPV + NPV}{2} \ \ \ \ \ \ \ \ \ \ Recall = \frac{TPR + TNR}{2} }
\end{equation}
Likewise, the $F_1$ Score (also known as the $F-Measure$) encapsulates the holistic predictive power of an algorithm with a focus on the minority class. Contemporary researchers including \cite{DalPozzolo2015} regard the  $F_1$ Score to be a suitable measure of predictability, even with imbalanced datasets. Much like AUPRC, the  $F_1$ Score combines  $PPV$ and $TPR$ to avoid bias introduced by disproportionately large $TN$ counts.

\begin{equation}\label{eq:F1Score}
{F_1 = 2\left(\frac{PPV * TPR}{PPV + TPR}\right) = \frac{2TP}{2TP + FP + FN} }
\end{equation}

%Cost function
Lastly, cost scores are generated using the framework outlined in Table 2. Since no company data is available for deriving explicit cost metrics, the cost of correct fraud detection $C_f$ is set at \$10, while  fraud error correction is set at $C_e$. The cost of not discovering fraud $C_l$ is also set at \$10. The fraud multiplier is set at \$2.40, cited from the 2016 LexisNexis Survey. The total cost of fraud $F_c$ for this study can be calculated as

\begin{equation*}
\label{eq:FraudCost}
{F_c = (TP + FN)C_{fl} + (FP)C_e + \sum_{i=1}^n{F_m(T_{ic} | T_{i = \textit{FN}}) - (T_{ic} | T_{i =\textit{TP}})}}
\end{equation*}
%Caveats about these fraud costs
where ${C_{fl} = C_l = C_f}$, $TP$ is the sum of true positives, $FP$ is the total of false positives,  and $T_i$ represents the $i^{th}$ transaction in a testing set of size $n$.  These variables are not based on empirical data in this study, and addressing a false positive is assumed to be less costly compared to addressing a genuinely fraudulent transaction. More importantly, this study is merely outlining a process by which companies could improve classification scoring. We recommend tailoring cost matrices to meet specific business needs.

%ASIDE FROM SOME EDITING, I AM READY FOR THE MODELS SECTION!!!!
% ADD FORMULAE TO THE GENETIC ALGORITHM
% Models subsection -- overview of the ones tested
\subsection{Implementing Practically Feasible Classifiers}

%Give an overview of each machine learning model used.
Acknowledging the challenges cited in \cite{Phua2010} of applying academic models to industry, we selected classifiers based on a combination of implementation practicality and past success detecting fraud. Higher-order profiling is impossible in this dataset, and its size inhibits the presence of concept drift. Therefore, these models strike a balance between time and spatial complexity as well as predictive accuracy.

\bigbreak
\textbf{Logistic Regression (LOG):} One of the most commonly implemented binary classification analyses \cite{hosmer2013applied}, Logistic Regression is a discriminant Bayesian model that approaches binary classification through direct calculation of Bayesian posterior $P(y|x)$ of the joint probability distribution $P(x,y)$ \cite{ng2002discriminative}. Logistic regressions assume little correlation (if any) exists between predictors in the feature space, as well as that few outliers and minimal skew are present. While characteristics of the provided data are not available in raw form, it can be assumed that PCA transformation approximately satisfies these requirements \cite{jolliffe1986principal}. Prevailing research also finds discriminative calculation -- even without computational or data quality considerations -- to be generally preferred \cite{ng2002discriminative}. The empirical findings of this model will be compared with Gaussian Naive Bayes predictors to test this hypothesis.

\bigbreak
\textbf{Linear Support Vector Classifier (SVC):} Vapnik (1995) \cite{Vapnik1995} first proposed Support Vector Machines (SVM) for pattern recognition by categorizing the problem as fitting an optimal separating hyperplane in $\mathbb{R}^n$ feature space, where $n$ is the number of features. By treating observations as $support \ vectors$, Vapnik characterizes the classifier as a Lagrangian optimization of the form:

\begin{equation}\label{eq:SVM}
{y(x) ={sign\Bigg[\sum_{i = 1}^N{\alpha_k y_k \Psi(x, x_k) + b \Bigg]}}}
\end{equation}
``for a training set of $N$ data points $\{y_k, x_k\}_{k=1}^N$  where $x^k \in \mathbb{R}^n$ is the $k^{th}$ input pattern and $y_k \in \mathbb{R}$ is the $k^{th}$ output pattern." \cite{Suykens1999}. $\alpha_k$ are positive, real constants, as is $b$. Notably, SVM makes use of kernel function $\Psi$, which transforms data to be optimal for hyperplane separation. Since Vapnik discovered Support Vector Machine's efficacy at pattern recognition, variant classifiers have become common due to the flexibility afforded by kernel functions.  For example, Suykens (1999) \cite{Suykens1999} determined a popular method for implementing SVM with least squares optimization. Kernel research continues to derive new applications for different disciplines. 

%How is SVC different
However, traditional Support Vector Machines are computationally expensive, even for Linear SVMs that utilize a kernel of the form $\Psi(x,x_k) = x_k^tx$.  To assuage resource constraints for classifying large datasets, \cite{ho2012large} developed a method for optimizing Linear Support Vector Classifiers without a kernel function. Instead, $L1$ and $L2$ regularization functions were employed.

%L1 regulatization function
\begin{equation}\label{eq:LAD}
{L1 ={argmin_w\sum_{i = 1}^n{\Bigg[y_i -  \sum_{j=0}^m{w_j x_{ij}\Bigg]^2 + \lambda\sum_{j=0}^m{|w_j|}}}}}
\end{equation}

%L2 regulatization function
\begin{equation}\label{eq:LSE}
{L2 ={argmin_w\sum_{i = 1}^n{\Bigg[y_i -  \sum_{j=0}^m{w_j x_{ij}\Bigg]^2 + \lambda\sum_{j=0}^m{w_j^2}}}}}
\end{equation}

%Add l1 and l2 loss functions here.
In machine learning, $L1$ loss is also known as Least Absolute Deviations (LAD) and included as part of regularization functions that classifiers seek to minimize. $L2$ is similar to $L1$ loss with the exception that it seeks to minimize the Least Squares Error (LSE). Both of these regularization functions are categorized for weights $w$, output label $y$, and prediction $x$ in equations \ref{eq:LAD} and \ref{eq:LSE}. $C$, another parameter tuned for Logistic Regression and Linear SVC, is a term for inverse regularization strength $\left(\frac{1}{\lambda}\right)$. This characterizes how harshly a model's complexity should be penalized during training.

\bigbreak
\textbf{Random Forests (RF):} Breiman (2001) \cite{Breiman2001} defines Random Forests as ``a combination of tree predictors such that each tree depends on the values of a random vector." An ensemble algorithm, Random Forest classifiers derive their efficiency from a sufficiently large number of decision trees. Each decision tree in a group may suffer from high generalization error and overfitting. But, if taken together as a random forest voting system, the ensemble has been proven to produce a "limiting ... generalization error" \cite{Breiman2001}. 

%Random forest performance when applied to fraud data
When applied as a fraud detection algorithm, Random Forests have also been deemed empirically robust against the challenges of data imbalance \cite{Bhattacharyya2011}. Moreover, Random Forest classification of credit card fraud was found by \cite{Whitrow2009} to be a sufficient means of classification, even with incomplete or imbalanced data. Random Forest's ensemble may be stochastically robust against the lack of information present in data employed by this study.

\bigbreak
\textbf{Gaussian Naive Bayes (GNB):} As opposed to the discriminant Bayesian implementation employed by Logistic Regression \cite{ng2002discriminative}, Gaussian Naive Bayes classification generates the joint probability distribution $P(x,y)$ through learning and then uses Bayes Theorem to calculate $P(y|x)$. Favored for its simple and efficienct implementation, Bayesian models consistently perform well across a variety of applications \cite{Kuo2004}. The high transaction volume for credit cards favors Bayesian models for their fast prediction generation \cite{Phua2010}, and many variant techniques like Bayes Minimum Risk (BMR) and Bayesian Neural Networks \cite{Stolfo2000} have found success in credit card fraud \cite{Bahnsen2016}. Outlined by \cite{Lewis1998}, Gaussian Naive Bayesian models assume features are independent and model a normal distribution. 

\begin{equation}\label{eq:NormBayesCalc}
{f(x \ | \ \mu, \sigma) = \frac{1}{\sqrt{2\pi}{\sigma}}{e}^{{-\frac{(x - \mu)^2}{2\sigma^2}}}}
\end{equation}
By using an empirically discovered $\mu$ and $\sigma$ for each feature, a Gaussian probability distribution can be derived, as shown in equation \ref{eq:NormBayesCalc}. This information is then be fed into record classification in equation \ref{eq:BayesianCalc}, where $P(A|B)$ is equal to the product of the probabilities of each feature $B_i$ belonging to the normal distribution of that feature $N(\mu_{B_i}, \sigma_{B_i})$ derived from training data.
\begin{equation}\label{eq:BayesianCalc}
{P(A | B) = P(A)  \prod_{i=1}^{n}P(B_i | A)}
\end{equation}

\bigbreak
\textbf{K-Nearest Neighbors (KNN):} The control classifier, K-Nearest Neighbors (KNN) is a rudimentary machine learning classifier. Though highly dependent on the dataset and application, KNN can be an effective means of prediction. A test prediction is generated by finding the $K$ most similar records in the training data, and returning the most common class label in the set. Similarity $d$ of a training record $x$ relative to test record $y$ is often determined using Euclidean distance (\ref{eq:KNN}) function defined for a dataset with $n$ features as

\begin{equation}\label{eq:KNN}
{d(x) = \sum_{i=1}^n{\sqrt{(x_i - y_i)^2}}}
\end{equation}

where $x_i$ is the $i^{th}$feature of the training record. KNN has been shown to be an effective algorithm for some classification problems, but it is commonly known \cite{Whitrow2009} to struggle with imbalanced data. Traditionally, an increase in the number of neighbors would correspond to a more general consideration of class characteristics and a better prediction. Imbalanced data prevents this improvement. As the number of neighbors increases, minority class representation in the $K$-subsample decreases, reducing algorithm performance.

\bigbreak
\textbf{Ensemble Classification and Genetic Algorithms}
Ensemble methods have found significant success across many fields of applied machine learning \cite{Bhattacharyya2011}. The ensemble implemented in this study takes the highest performing individual classifiers and weights them to produce a classification in the form of a weighted average. Ensemble studies in outlier detection and other disciplines \cite{aggarwal2001outlier} have determined genetic algorithms to be a suitable method for finding optimal parameters in high dimensional feature space.

%How are weights handled by the algorithm
The algorithm begins with an array of random weights, one for each algorithm. These weights are chosen from the uniform distribution $U \sim [1,2^{40})$ so integer weights are defined by 40 bits or fewer. This allows the genetic algorithm to treat the binary representation of these numbers as "genes" and vary them throughout the evolution process. For each generation, 50 arrays are generated, each with a probability of succession $P(s)$ determined by its fitness $f$. To determine this probability, fitness measurements (fraud costs) are first converted to positive values $f_p$ by the transformation in equation \ref{eq:positive weights}

\begin{equation}\label{eq:positive weights}
{f_{p_i} = {f_i + f_{min} + 1}}
\end{equation}
where $f_{pi}$ is the positive fitness of the $i^{th}$ weight array and  $f_{min}$ is the lowest fitness in the generation. It is important to note that since cost is used as a fitness metric, low fitness scores are desired. Therefore,  $f_p$ values are inverted before probabilities are taken. This calculation takes the form of equation \ref{eq:probgenetic}

\begin{equation}\label{eq:probgenetic}
{P(x) = \frac{f_{p_{max}} + 1 - f_{p_i} }{\sum_{i=1}^n{(f_{p_{max}} + 1 - f_{p_i})}}}
\end{equation}
where $f_{p_{max}}$ is the maximum (worst) fitness score. In turn, all of the weights in the array sum to one. With these probability scores, the following generations are populated by repeatedly choosing two weight arrays from the existing population and generating a child array with bit-manipulation crossover. Bit flipping is also used to randomly mutate arrays in a population, though this is set to occur rarely. To prevent the ensemble from over-weighting an individual model, a ceiling weight $w_{max}$ is set at 0.49. Each ensemble optimization via genetic algorithm is given a specific runtime of 1 minute. 
%%%%%%%%%%%%%%%%%%%%%%%%%%%%%%%%%%%%%%%%%%%%%%%%%%%%%%%%%%
%
% Results and Discussion Sections
% 
%%%%%%%%%%%%%%%%%%%%%%%%%%%%%%%%%%%%%%%%%%%%%%%%%%%%%%%%%%%%%%%%%%%%%%%%%%%%%%%%%%%%%%%%%%%

\section{Empirical Findings}
\label{S:8}

\subsection{Undersampling generates superior model performance relative to SMOTE}

%Overview of the 
Sample size and fraud ratio combinations are listed in Table 3, with the best sample parameters for each method noted as well. Since undersampling did not depend on a set sample size, we were able to computationally search the entire sample-model feature space. That is, for each fraud ratio derived undersample, every model was tuned and tested for all parameter combinations listed in Table 4. 

% Sampling Tests
\begin{table}[h]
    \centering
    \caption{Tested Sampling Combinations}
    \begin{tabular}{
    	l
        l
        l
        l
        }
        \toprule
{Sample}& {Sample}& {Fraud}& {Best Sample}\\
{Method} & {Size $(Thous.)$} & {Ratio} & {$(size, ratio)$} \\
        \midrule
		SMOTE & 1, 2, 3, 5, 10 & 0.1, 0.2, ... 0.5 & (1, 0.5) \\
        Under & - & 0.1, 0.2, ... 0.5 & ( - , 0.3)\\
        \bottomrule
    \end{tabular}
\end{table}

%How SMOTE was optimized with Bootstrapping
By contrast, SMOTE's dependency on both a specific sample size and fraud ratio made it computationally infeasible to search the entire model-sample feature space outlined in Table 4. Therefore, a bootstrapping method was implemented to indirectly converge on the optimal sample size and fraud ratio. First, all samples were tested with default parameters assigned to each model. Default parameters are provided by Sci-kit Learn, and are listed in Table 5. After testing all SMOTE sample sizes and fraud ratio combinations, the sample with the lowest average cost across all models is discerned and used to test all parameters. With sampling parameters held constant, optimal parameters are identified independently for each algorithm. 

\begin{figure}[!ht]
\label{fig:SMOTEUnderComp}
\centering\includegraphics[width=1\linewidth]{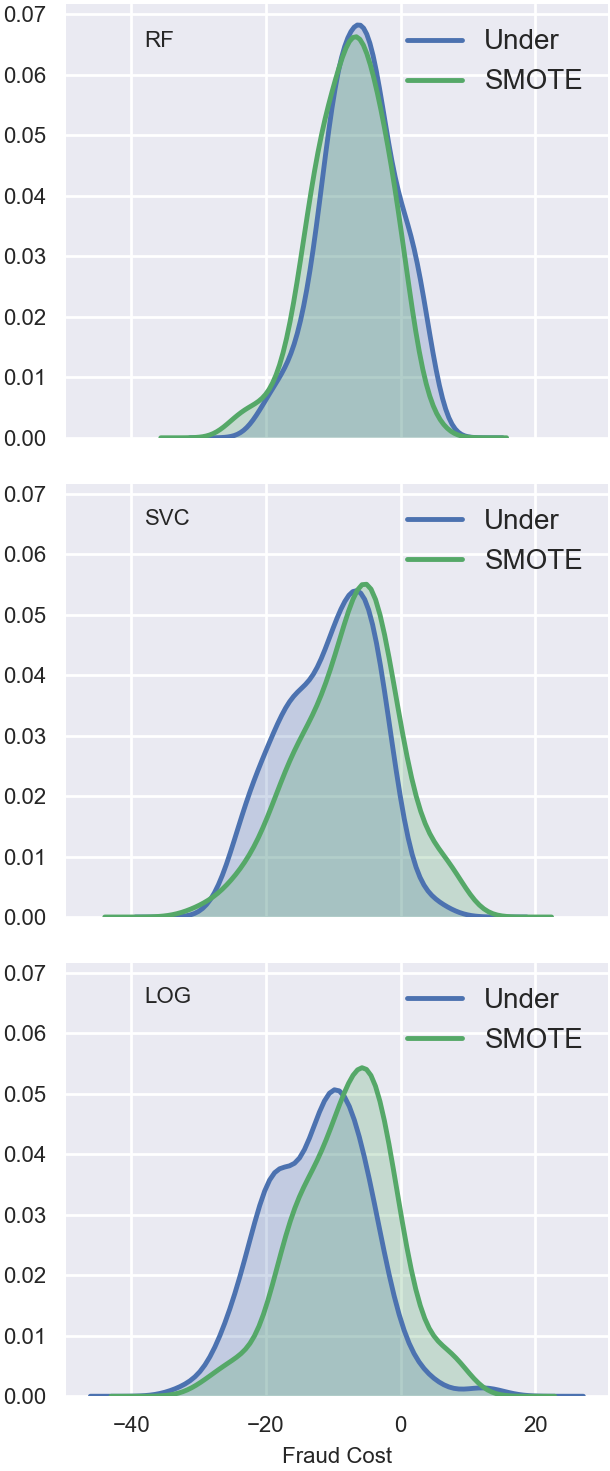}
\caption{Comparison of SMOTE and Undersampling methods as they pertain to training RF, SVC, LOG classifiers to minimize fraud cost. All figures are cumulative frequency histograms. Fraud cost given in USD - Thousands.}
\end{figure}

%Bootstrapping continued
Classifier parameters are then fed back in to the model and held constant as sample-ratio combinations are tested again. This process continues until empirical results converge on a single sample-ratio combination. Random model parameters and sample-ratio combinations were also tested to independently verify a local optimum. If independent findings were inconsistent with the current results, bootstrapping continued using the random parameters aforementioned. All iterations ran 100 Monte Carlo simulations for every unique sample-ratio-model combination.

% Total tested Parameters
\begin{table}[h]
    \centering
    \caption{Model Tuning Tested Parameters}
    \begin{tabular}{
    	l
        l
        l
        l
        l
        }
        \toprule
        \multicolumn{1}{c}{} &
        \multicolumn{4}{c}{Parameters}\\
        \cmidrule(lr){2-5} 
{Model}& {Penalty}& $C$& {Trees $(RF)$}\\
{Name} &  &  & {or$\ K \ (KNN)$}\\
        \midrule
		LOG& l1, l2  & 0.5, 1, 5, 10, 20 & &\\
        SVC & l1, l2  & 0.5, 1, 5, 10, 20 & &\\
        RF &  && 10, 20 ... 100\\
        KNN &  &&10, 20 ... 100\\
        \bottomrule
    \end{tabular}
\end{table}

%Talk more about bootstrapping
While searching for the ideal SMOTE sample, models with outlier performance were detected and dropped. For a classifier to be dropped from the remainder of the study, its fraud cost had to be significantly higher than the average of all models in both undersampling and SMOTE samples.

%KNN dropped as well as L1 and L2 loss functions
As noted in Tables 5 and 6, the fraud cost of the K-Nearest Neighbors algorithm was orders of magnitude higher than other models, regardless of parameter tuning. Therefore, its performance was not considered during the bootstrapping process. Gaussian Naive Bayes, though lower in cost than KNN, was also dropped for the same reason (see Figure 3). Additionally, Logistic Regression and Linear SVC Classifiers optimized with $L2$ regularization functions were dropped as the difference in performance between $L1$ and $L2$ regularized functions was significantly large (see Figures 6 and 7). 

%Talk more about what was significant for SMOTE sampling
In general, a SMOTE samples fraud ratio was the most indicative of its ability to train highly efficient, low cost algorithms. Though fraud costs varied by iteration, most algorithms experienced improved performance as the  fraud ratio increased up to $50\%$. After a threshold of 30\%, fraud cost differences in many models were not significant enough to be distinguished from variance in performance. In the same vein, many models (Table 6) depicted a slight positive relationship between rising cost and rising sample size.

%Trends for undersampling
Undersampling did not possess a consistent trend between fraud ratio and model performance. Gaussian Naive Bayes and Random Forest Classifiers continuously improved as the fraud ratio increased. Conversely, regression-based models Linear SVC and Logistic classification possessed a quadratic relationship with a fraud ratio that peaked at $30\%$, decreasing thereafter. Logistic Regression and Linear SVC also had higher average performance for a fraud ratio of 0.2, but this was overshadowed by Random Forest's increase in performance as it moved from a fraud ratio of 0.2 to 0.3.  Repeatedly sampling verifies all of these trends with relative certainty despite the stochastic nature of sampling methods used in this analysis. Thus, the optimal fraud ratio was set a 0.3 to embody the best average performance for the three best classifiers. Using the best holistic sample, model parameters were tuned to minimize the cost of fraud.

%Raw Model Performance Change name of Figure
\FloatBarrier
\begin{figure*}[!h]
\label{fig:RawPerf}
\centering
\includegraphics[width=1\linewidth]{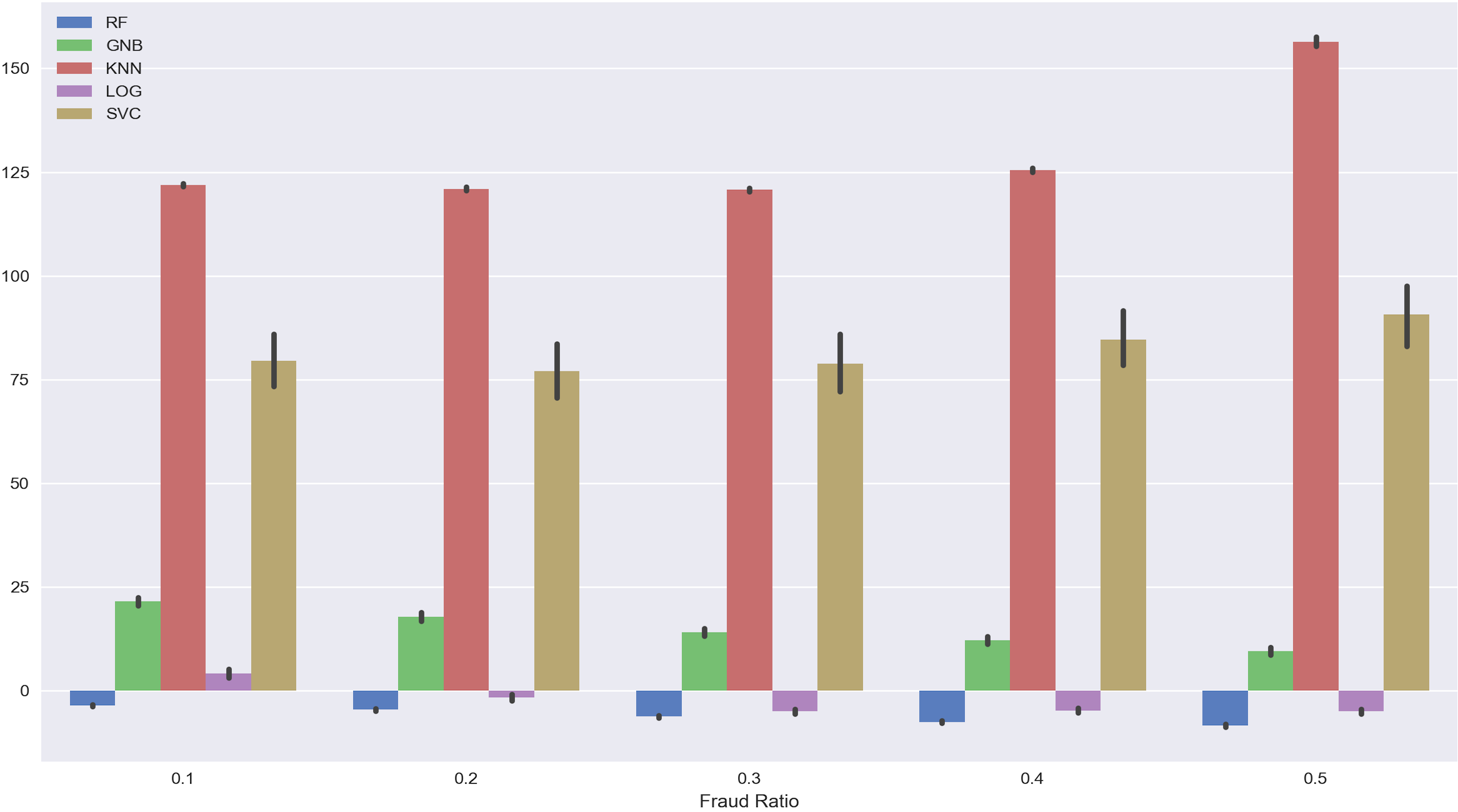}
\caption{Mean performance of all FDSs, all parameters. The y-axis represents fraud cost in USD - Thousands, and black lines signify sample $\sigma$.}
\end{figure*}
\FloatBarrier

\subsection{Optimal model parameters consistent across sample types}
 
Optimal size and fraud ratios for SMOTE and undersampling were held constant as different parameters for Fraud Detection Systems were tested. With the exception of Random Forest Classifiers, no algorithm proved to be robust across all of its parameter combinations. 

% BEST and default Parameters
\begin{table}[!h]
    \centering
    \caption{Parameter Legend}
    \begin{tabular}{
    	l
        l
        l
        l
        }
        \toprule
        \multicolumn{1}{c}{} &
        \multicolumn{2}{c}{SMOTE} &
        \multicolumn{1}{c}{Under} \\
        \cmidrule(lr){2-3} 
        \cmidrule(lr){4-4} 
{Model}& {Default}& {Best.}& {Best.}\\
{$(Params.)$} & {Param.} & {Param.} & {Param.} \\
        \midrule
		LOG\\
        $(Pen. \ C)$& l2, 1 & l1, 0.5 & l1, 0.5\\\\
        SVC \\
        $(Pen. \ C)$& l2, 1 & l1, 0.5 & l1, 0.5\\\\
        RF \\
        $(\# \ Trees)$& 10 & 80 & 80\\\\
        KNN \\
        $(K)$& 5 & 10 & 10\\
        \bottomrule
    \end{tabular}
\end{table}

%KNN performance (dropped)
K-Nearest Neighbors achieved its highest predictive power with a $K$ of 10. Further increases in the number of neighbors gradually decreased the $F_1$ Score and increased the fraud cost of the algorithm. While the root cause of this trend is unknown, it is possible that the minority-majority class boundary in the training data is not well defined. KNN's predictive accuracy is dependent on a well-defined boundary between majority and minority data. Credit card fraud detection can be categorized as an outlier detection problem, and therefore may not have enough minority class information to generalize a boundary in the training data. Concept drift exacerbates this issue, potentially causing minority records in training data to not be representative of the testing sample. Thus, KNN was dropped as a suitable fraud detection system.

%Gaussian Naive Bayes Performance
Gaussian Naive Bayes does not have parameters with which it may be tuned. Consistently, the fraud cost of GNB decreases as the fraud ratio of the sample data increases. Even so, data outlined in Table 6 denotes GNB was unable to consistently generate an average fraud cost below \$0 for undersamples. With no additional tuning possible, the GNB Fraud System is impossible to improve. Therefore, it was also dropped from the analysis. 

%Random forest regression
Random Forest classifiers performed best in situations where the minority class was represented most in sample. Cost savings for these Fraud Detection Systems was the lowest for samples with a fraud ratio of $50\%$ across both undersampling and SMOTE. Also, models optimized best with the smallest SMOTE samples with $1000$ records, holding fraud ratios constant. Perhaps decision trees built primarily on synthetic records are over-fitting to the training dataset and unable to generalize rules for identifying fraud. Synthetic records do not greatly deviate from the originals from which they were derived and may not allow classification boundaries to generalize.

%%%%%%%%%%%%%%%%%%%%%Tables for sampling%%%%%%%%%%%%%%%%%%%%%%%
%Under sample tuning
\FloatBarrier
%Under sampling
\begin{table*}[t]
    \centering
    \caption{Under Sample Tuning Summary: Best Parameters}
    \begin{tabular}{
        S[table-format = 0.1]
        S[table-format = 2.2]
        S[table-format = -5.0]
        S[table-format = 2.2]
        S[table-format = -5.0]
        S[table-format = 2.2]
        S[table-format = -4.0]
        S[table-format = 2.2]
        S[table-format = 5.0]
        S[table-format = 1.2]
        S[table-format = 6.0]
        }
        \toprule
        \multicolumn{1}{c}{} &
        \multicolumn{10}{c}{Model Performance}\\
        \cmidrule(lr){2-11} 
{Fraud}& {LOG}& {LOG} & {SVC} & {SVC} & {RF} &{RF} & {GNB} & {GNB}& {KNN} & {KNN} \\
{Ratio} & { $F_1 \ (\%)$} & {Cost $(\$)$} & { $F_1 \ (\%)$} & {Cost $(\$)$} & { $F_1 \ (\%)$} & {Cost $(\$)$} & { $F_1 \ (\%)$} & {Cost $(\$)$} & { $F_1 \ (\%)$} & {Cost $(\$)$} \\
        \midrule
          0.1& 33.50	&	-4992	&	36.95	&	-4392	&	53.80	&	-3898	&	17.08	&	21513	&	2.01	&	116711	\\
0.2 &	18.28	&	-9134	&	17.05	&	-8916	&	32.24	&	-5214	&	15.78	&	17897	&	1.28	&	111311	\\
0.3 &	11.60	&	-12318	&	9.77	&	-10744	&	19.43	&	-6365	&	14.41	&	14082	&	0.87	&	113065	\\
0.4	&7.62	&	-11011	&	6.00	&	-9520	&	12.49	&	-7969	&	13.56	&	12159	&	0.61	&	121524	\\
0.5&	4.94	&	-8888	&	3.95	&	-5942	&	8.93	&	-8894	&	12.37	&	9508	&	0.46	&	135647	\\
	
        \bottomrule
    \end{tabular}
\end{table*}
\FloatBarrier

%%%%%%%%%%%%%%%%%%%%%%%%%%%%%%%%%%%%%%%%%%%%%%%%%%%%%%%%%%%%%%%%%%%%

%SMOTE sampling
\FloatBarrier
\begin{table*}[t]
    \centering
    \caption{SMOTE Sample Tuning Summary: Best Parameters}
    \begin{tabular}{
    	l
        S[table-format = 0.1]
        S[table-format = 2.2]
        S[table-format = -5.0]
        S[table-format = 2.2]
        S[table-format = -4.0]
        S[table-format = 2.2]
        S[table-format = -4.0]
        }
        \toprule
        \multicolumn{2}{c}{} &
        \multicolumn{6}{c}{Model Performance}\\
        \cmidrule(lr){3-8} 
{Sample}& {Fraud}& {LOG}& {LOG} & {SVC} & {SVC} & {RF} &{RF} \\
{Size} & {Ratio} & { $F_1 \ (\%)$} & {Cost $(\$)$} & { $F_1 \ (\%)$} & {Cost $(\$)$} & { $F_1 \ (\%)$} & {Cost $(\$)$} \\
        \midrule
          1,000	&	0.2	&	21.40	&	-8416	&	21.59	&	-6931	&	40.94	&	-4738	\\
	&	0.3	&	15.88	&	-9286	&	15.43	&	-9289	&	29.58	&	-4663	\\
	&	0.4	&	11.80	&	-10685	&	11.78	&	-9499	&	25.05	&	-7329	\\
	&	0.5	&	8.49	&	-10012	&	8.22	&	-9681	&	18.76	&	-8290	\\\\
2,000	&	0.1	&	38.82	&	-5571	&	44.05	&	-4587	&	50.80	&	-4024	\\
	&	0.2	&	23.65	&	-8671	&	25.19	&	-6925	&	44.39	&	-5898	\\
	&	0.3	&	18.83	&	-7854	&	20.09	&	-7249	&	38.30	&	-5492	\\
	&	0.4	&	12.57	&	-7979	&	12.94	&	-7471	&	32.56	&	-5706	\\
	&	0.5	&	9.32	&	-9331	&	9.61	&	-9157	&	26.83	&	-7708	\\\\
3,000	&	0.1	&	39.65	&	-4672	&	46.36	&	-3463	&	52.71	&	-4410	\\
	&	0.2	&	25.48	&	-6647	&	26.97	&	-5555	&	49.19	&	-4894	\\
	&	0.3	&	18.88	&	-8997	&	20.10	&	-7866	&	42.20	&	-5726	\\
	&	0.4	&	15.12	&	-8690	&	15.68	&	-7935	&	39.49	&	-5036	\\
	&	0.5	&	10.59	&	-8499	&	10.77	&	-7333	&	32.85	&	-6091	\\\\
5,000	&	0.1	&	43.43	&	-3039	&	51.36	&	-2358	&	56.17	&	-4218	\\
	&	0.2	&	27.79	&	-6011	&	29.79	&	-4831	&	52.83	&	-3749	\\
	&	0.3	&	20.09	&	-6956	&	21.71	&	-6391	&	50.00	&	-4366	\\
	&	0.4	&	15.05	&	-7907	&	15.91	&	-7233	&	46.10	&	-4805	\\
	&	0.5	&	11.26	&	-8008	&	11.71	&	-7098	&	42.54	&	-4671	\\\\
10,000	&	0.1	&	43.25	&	-4299	&	51.76	&	-3172	&	58.21	&	-4225	\\
	&	0.2	&	28.53	&	-7590	&	31.20	&	-6424	&	56.17	&	-5368	\\
	&	0.3	&	21.56	&	-7575	&	23.30	&	-7080	&	55.15	&	-4191	\\
	&	0.4	&	16.22	&	-8727	&	17.25	&	-8035	&	52.81	&	-4147	\\
	&	0.5	&	12.97	&	-9276	&	13.43	&	-8236	&	50.71	&	-4740	\\
	
        \bottomrule
    \end{tabular}
\end{table*}
\FloatBarrier

%%%%%%%%%%%%%%%%%%%%%%%%%%%%%%%%%%%%%%%%%%%%%%%%%%%%
\clearpage
\pagebreak
\newpage
%Random Forests
Regardless, Random Forest Classifiers trained on both samples performed optimally with the maximum of $100$ decision trees. Larger tree counts may have performed even better, but experimentation was practically limited by sample size constraints. An undersample with a $50\%$ fraud ratio may only contain 160 records.

%Evolution of RF performance as shown on Figure 4
Outlined in Figure 4, Random Forest detection of fraud is particularly robust. Average fraud cost starkly shifts as models went from 10 to 20 trees, as did fraud cost variation. Cost continued to drop as more trees were added, but only slightly compared to initial improvements. Near 90 and 100 neighbors, changes in mean performance are too minute to differentiate from sample variation. However, it appears that the distribution skews left as more trees are added. As expected, Random Forest classification was one of the most robust models to concept drift, data imbalance, and even its own meta-parameters.

%Talk about Logistic Regression
Logistic Regression was the most efficient model at minimizing the cost of fraud. Each classifier tested was tuned by two parameters: a regularization function ($penalty$) and the strength term $C$. A scalar coefficient that calibrates regularization strength (see Table 4), fraud detection models worked best with a low C value, or a high penalty $\lambda$ for regularization. Contextually, a high $\lambda$ value may be necessary to create a functional FDS. Little fraud data is available, leading many alternative samples to be small. Low regularization penalties may quickly lead to over-fitting and generalization error. 

%Linear SVC performance with C
Cost-based performance with Linear SVC algorithms also improves as regularization strength $\lambda$ increases. Fraud cost trends witnessed during model tuning for Logistic Regression are directly paralleled by Linear SVC. As noted in the next section, both were heavily influenced by different regularization functions.

\subsection{Regularization central to LOG and SVC performance}

%Notes about what is used to optimize the graphs
Articulated by equations \ref{eq:LAD} and \ref{eq:LSE}, two regularization functions were utilized to fit both Logistic and Linear SVC classifiers to fraud data. While  implementations are similar, resulting fraud costs for each type were quite different. These differences are visualized with the illustrations in Figure 6 and 7 for Linear SVC and Logistic Regression, respectively. 

%Notes about the graphs
Mapped on a feature space of $F_1$ performance and fraud cost, a performance distribution. The darkness of each section correlates to the density of observations found therein. Also note that each subplot in Figures 6 and 7 has a unique $y-range$, though all share the same $x-axis$. The purpose of this is to compare the $F_1$ scores while simultaneously underscoring the stark differences in fraud cost between $L1$ and $L2$ optimized Fraud Detection Systems.

%Show the differences in Linear SVC
Figures 3 and 5 for articulate the stark impact regularization had on performance for Linear SVC classification. Across all regularization strength parameters, Linear SVC classifiers optimized with Least Square Error regularization had an average fraud cost of \$169,694 compared to a cost of -\$5196 for Least Absolute Deviation, a difference of \$174,890 over two days. Annually, the difference in cost savings for these classifiers would amount to millions, highlighting the central importance of model tuning.

\FloatBarrier
\begin{figure}[!ht]
\label{fig:TunedPerf}
\centering 
\includegraphics[width=1\linewidth]{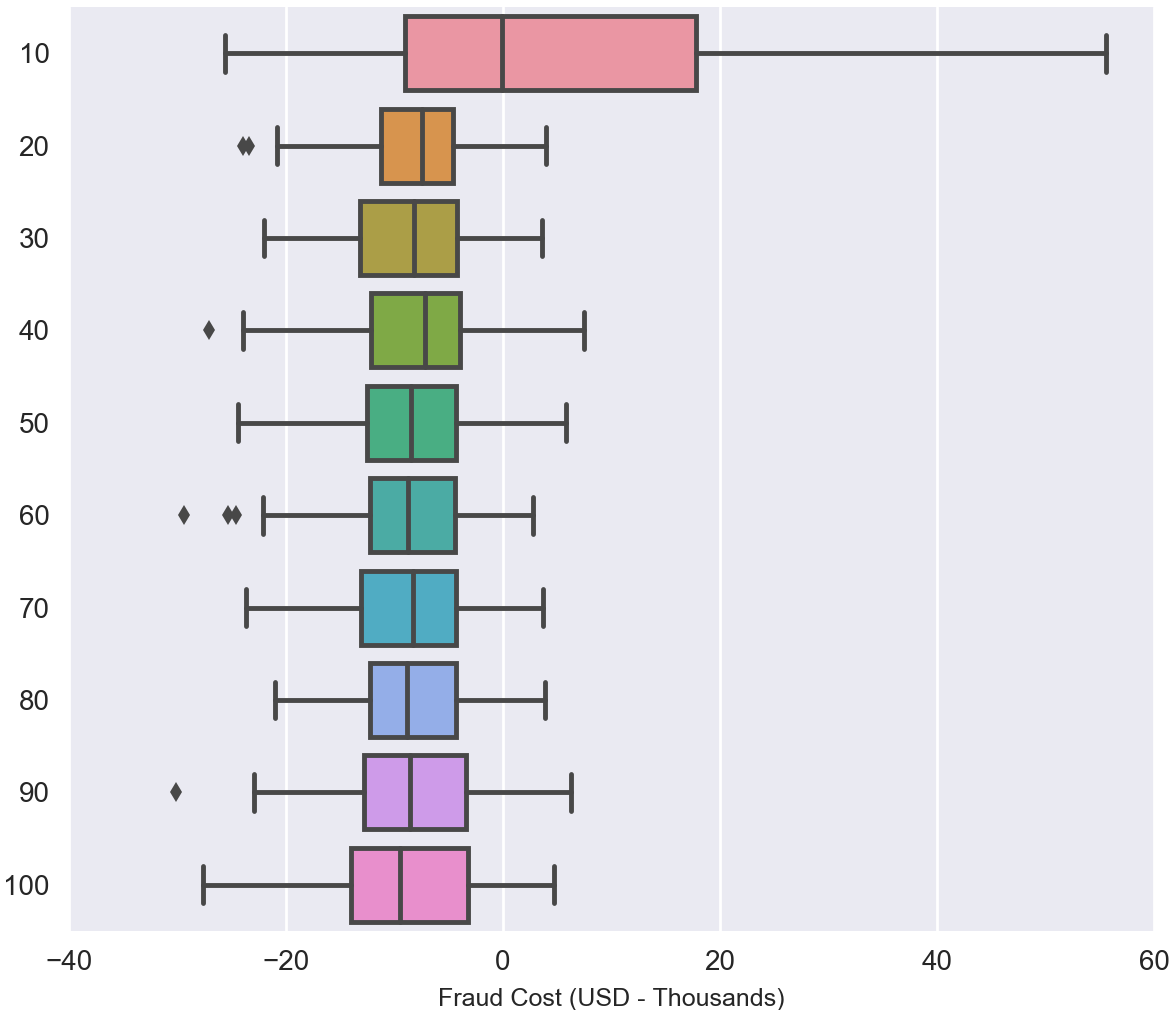}
\caption{Fraud cost of RF classification using different numbers of decision trees ($y-axis$) in ensemble. Fraud cost represented in USD - Thousands.}
 \vspace*{\floatsep}
  \vspace*{\floatsep}
\includegraphics[width=1\linewidth]{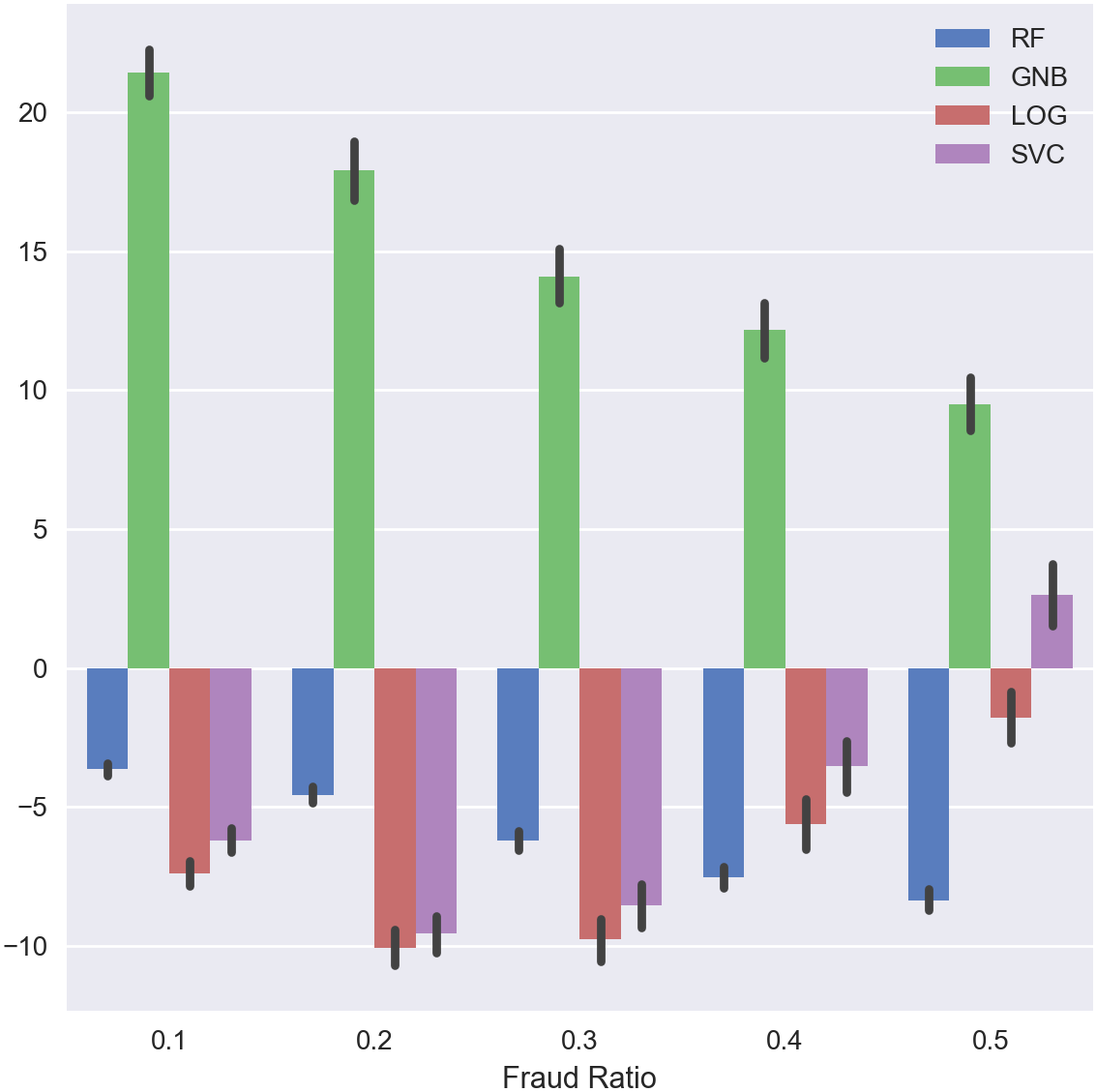}
\caption{Model performance without outliers. LOG and SVC models trained with $L2$ regularization have been dropped. The y-axis represents fraud cost in USD - Thousands, and black lines signify sample $\sigma$.}
\end{figure}
\FloatBarrier

Figure 6 provides further detail on Linear SVC cost discrepancies. The consistency of performance for $L1$ models is depicted by the unimodal distribution around a cost of -\$10,000 and a $F_1$ Score of 0.1. Conversely, the same models tuned with $L2$ loss possess a polymodal distribution centered on an $F_1$ Score of approximately 0.03. The $L2$ trained performance distribution appears to achieve costs in excess of \$200,000 fairly consistently.

\begin{figure}[!ht]
\label{fig:TunedPerf}
\centering 
\includegraphics[width=1\linewidth]{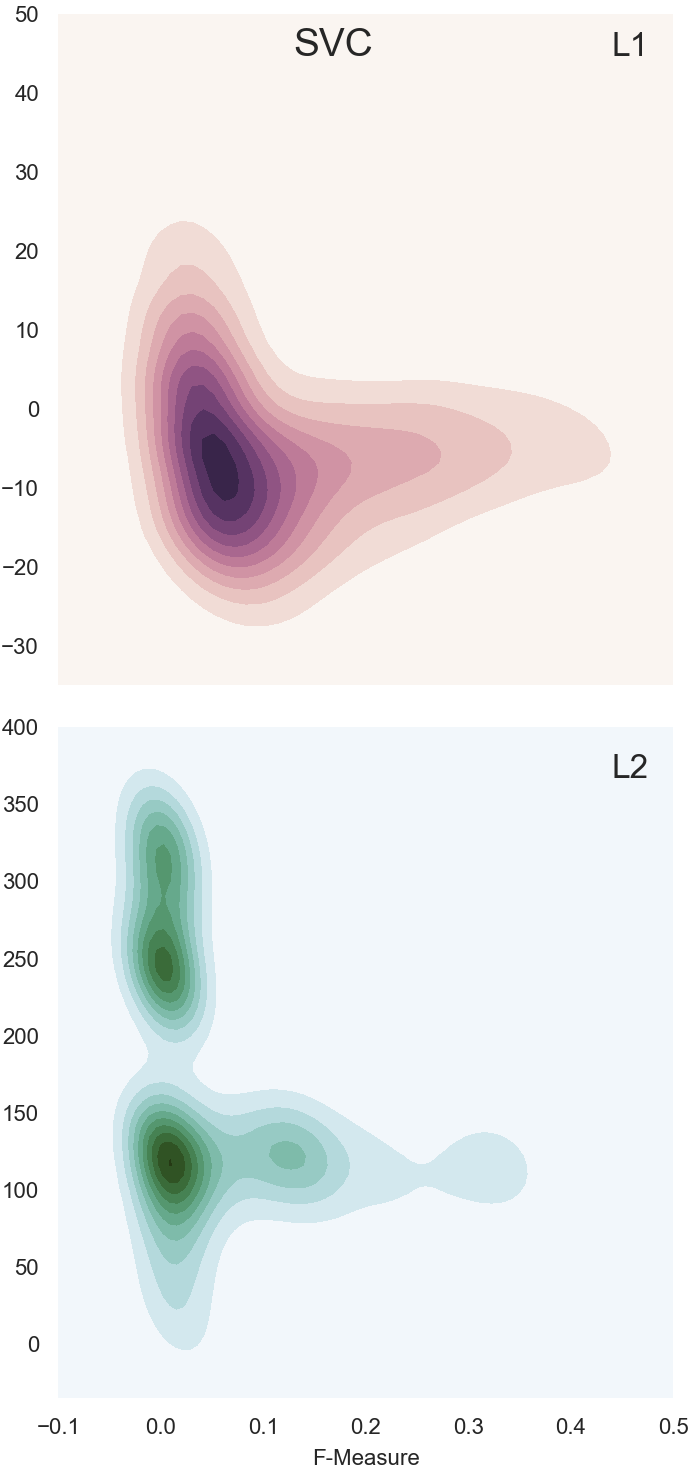}
\caption{Linear SVC comparison of L1 ($LAD$) and L2 ($LSE$) regularization functions. The $y-axis$ represents fraud cost in USD - thousands.}

\end{figure}

\begin{figure}[!ht]
\label{fig:TunedPerf}
\centering 
\includegraphics[width=1\linewidth]{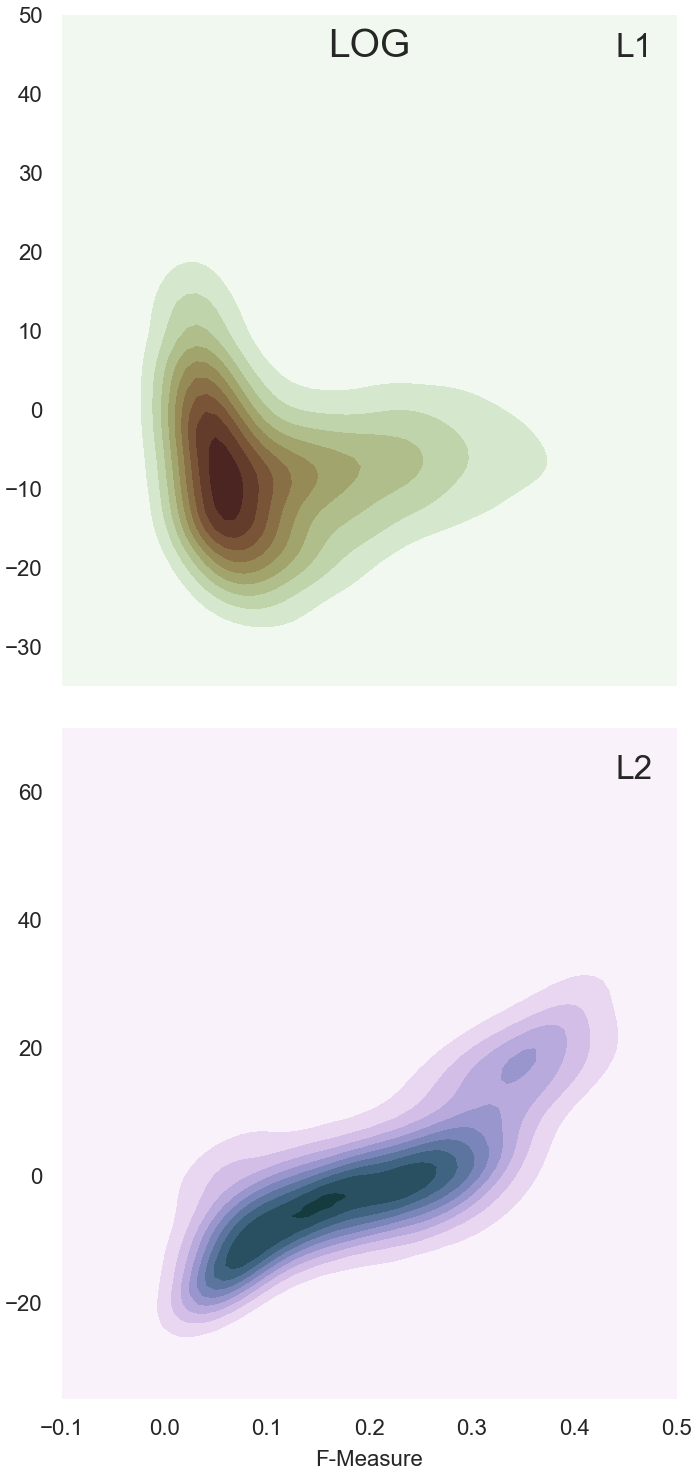}
\caption{Logistic Regression comparison of L1 ($LAD$) and L2 ($LSE$) regularization functions. The $y-axis$ represents fraud cost in USD - thousands.}

\end{figure}

%%%%%%%%%%%%%%%%%%%%%%%%%%%%%%%%%%%%%%%%%%%%

%Information about logistic regression now
Logistic regression maintains the same cost differences between regularization practices, though the effect is far less pronounced. On average, the performance of $L2$ models across all $C$ parameters is approximately equal to \$2521. Though a respectable fraud cost, $L1$ trained algorithms generate an average cost of \$6988 under the same circumstances. Additionally, Figure 7 describes the differences in performance distributions for Logistic Regression on both traditional and cost-based metrics.

%Difference in logistic metrics based on Figure 7
Again, performance distribution for $L2$ trained models is much less concentrated for Logistic Regression. Trending in the plot seems to indicate a slight positive relationship between $F_1$ Score and higher cost. This positive correlation contradicts the notion that a higher $F_1$ Score should correspond to lower cost. Visually, the SVC performance distribution appears nearly identical to that of Logistic Regression when trained with $L1$ regularization, implying $L1$ is effective for imbalanced data.

\clearpage
\pagebreak
\newpage
\FloatBarrier
\begin{figure}[!ht]
\label{fig:TunedPerf}
\centering 
\includegraphics[width=0.98\linewidth]{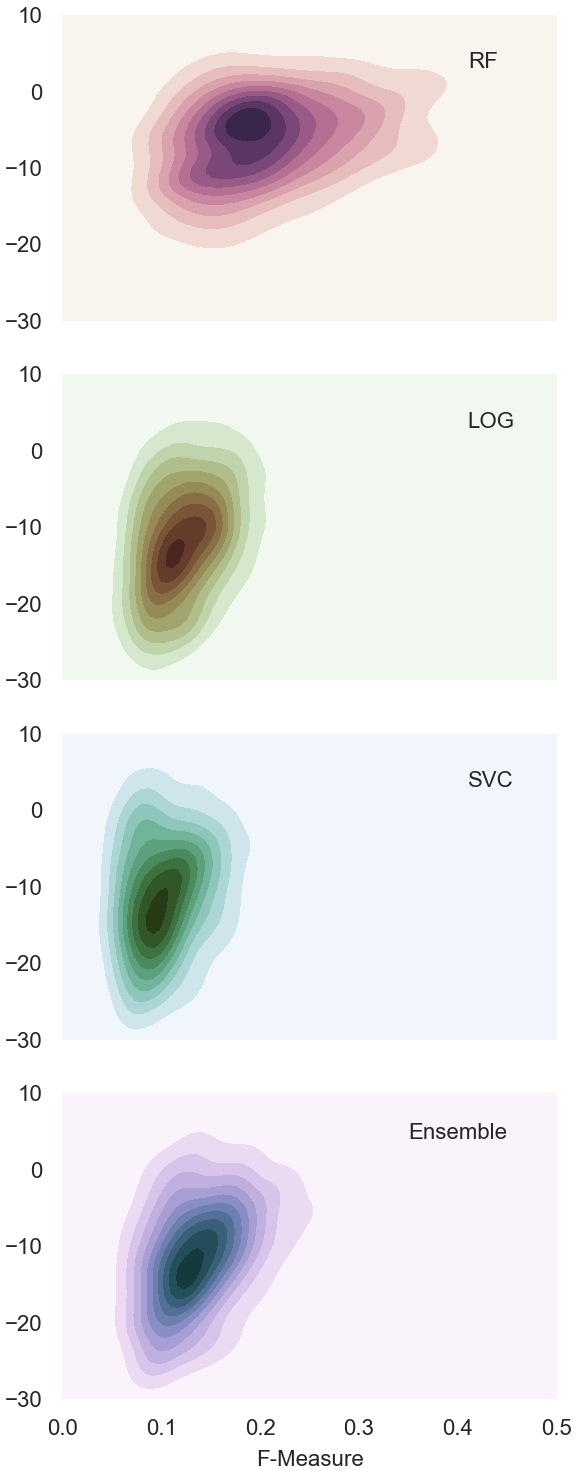}
\caption{Ensemble performance relative to its components based on $F_1$ score and fraud cost. All models have identical axes, and the $y-axis$ represents fraud cost in USD - thousands.}
\end{figure}
\FloatBarrier

\subsection{Ensemble unable to outperform individual components.}

%Overview of the ensemble
The top three performing models by fraud cost: Random Forest, Linear SVC, and Logistic Regression, were chosen as the components of the ensemble. Due to the stochastic nature of genetic algorithms, ensemble performance, and random sampling, Monte Carlo tests for ensemble efficacy were run 1000 times.  For the majority of these iterations, at least one weight converged to the ceiling threshold of $49\%$.

%Explain why undersampling used
Undersampling was yielded the most cost-effective Fraud Detection Systems, and therefore it was used to test ensemble performance. Undersamples with a fraud ratio of 0.3 were taken an then split into 60\% and 40\% training and testing sets, respectively. Optimal model parameters (Table 5) were also utilized. Interestingly, these subsets were roughly comprised of 100 records each, an incredibly small percentage of the sampling population. Yet, as depicted by Figure 8, all algorithms were able to achieve costs far lower than those built with larger samples.

%What were the results of running ensemble
In particular, Random Forest proved to be the most consistent model regarding fraud cost, but the most variable regarding $F_1$ scores as denoted by the width of its distribution plot. Logistic and Linear SVC classifiers were far more variable in their cost output despite consistent $F_1$ scores. Ensemble performance in the bottom plot embodies both of these distributions, but was unable to attain a lower average cost performance that its components.

%Explanation of the samples 
With an average fraud cost of -\$10,847, the ensemble algorithm outperformed all other models except Logistic Regression, which had an mean cost of -\$11,168. Yet, it had a lower cost variance than Logistic Regression and Linear SVC, as seen in Figure 9. Though sub-optimal when considered solely for its cost saving potential, ensemble consistency may be particularly valuable if high variation is present in existing transaction activity. Striking a balance between performance and variation is central to long-term fraud detection success.

\FloatBarrier
\begin{figure}[!ht]
\label{fig:TunedPerf}
\centering 
\includegraphics[width=1\linewidth]{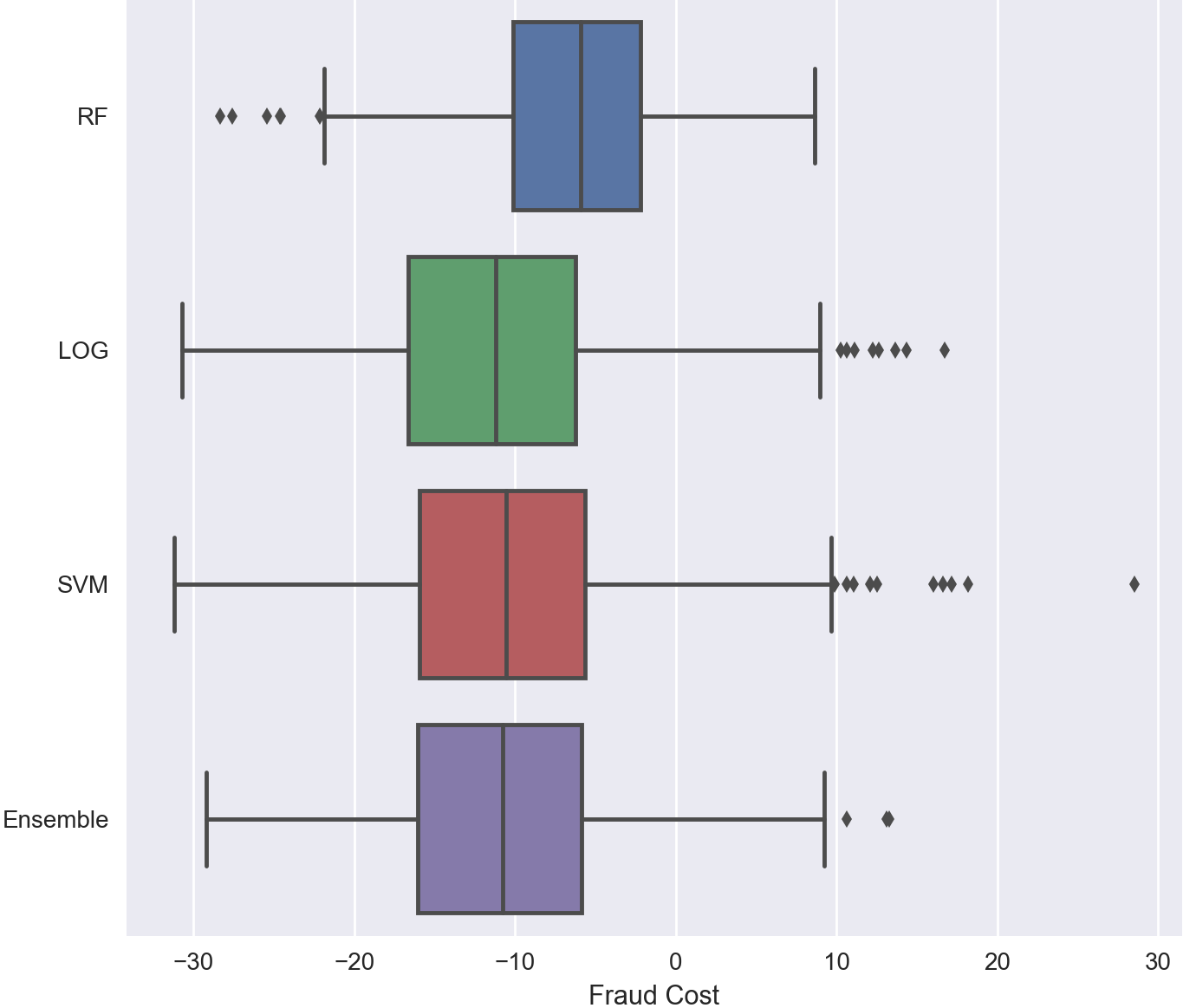}
\caption{Box plot of ensemble and component fraud cost (USD - thousands)}

\end{figure}
\FloatBarrier

\begin{table*}[t]
    \centering
    \caption{Ensemble Performance Relative to Components}
    \begin{tabular}{
        l
        S[table-format = 3]
        S[table-format = 2]
        S[table-format = 6.0]
        S[table-format = 4.0]
        S[table-format = 2.2]
        S[table-format = 2.2]
        S[table-format = 2.2]
        S[table-format = 2.2]
        S[table-format = 2.2]
        S[table-format = 2.2]
        S[table-format = 2.2]
        S[table-format = 2.2]
        S[table-format = -5.0]
        }
        \toprule
        \multicolumn{4}{r}{Fraud Counts} & 
        \multicolumn{7}{r}{Model Performance}\\
        \cmidrule(lr){2-5}
        \cmidrule(lr){6-14} 
        {Model}&{TP} & {FN}& {TN} & {FP} & {$TPR$} & {$TNR$} &{$PPV$} & {$NPV$} & {Rec.}& {Prec.} & {Acc.} &{$F_1$}&Cost \\
        {$(Param.)$} & {$(\#)$} & {$(\#)$} & {$(\#)$} & {$(\#)$} & {$(\%)$} & {$(\%)$} & {$(\%)$} & {$(\%)$} & {$(\%)$} & {$(\%)$} & {$(\%)$} & {$(\%)$}& {$(\$)$} \\
        \midrule

\textbf{SVC}	\\														
$(Pen., C)$	\\																			
l1, 0.5	&	348	&	45	&	277542	&	6544	&	88.50	&	97.70	&	5.05	&	99.98	&	93.10	&	52.52	&	97.68	&	9.56	&	-10548	\\\\
																											
\textbf{RF}	\\																			
$(Trees)$	\\																		
80	&	341	&	53	&	281175	&	2911	&	86.61	&	98.98	&	10.48	&	99.98	&	92.79	&	55.23	&	98.96	&	18.70	&	-6353	\\\\
																											
\textbf{LOG}	\\																	
$(Pen., C)$	\\																		
l1, 0.5	&	349	&	45	&	278807	&	5278	&	88.60	&	98.14	&	6.20	&	99.98	&	93.37	&	53.09	&	98.13	&	11.58	&	-11168	\\\\
																					
\textbf{Ensemble}	&	348	&	46	&	279411	&	4675	&	88.36	&	98.35	&	6.92	&	99.98	&	93.36	&	53.45	&	98.34	&	12.84	&	-10847	\\
        \bottomrule
    \end{tabular}
\end{table*}
\FloatBarrier

%Talk about how other methods may be able to yield good results as well
More sophisticated ensemble classifiers may take greater advantage of the balance between low fraud cost and consistent performance. Avenues for further research may include refining the ensemble optimization process or assigned specific models to detect fraud and non-fraud transactions. Currently, all models seek to accomplish the same goal of preventing fraud by categorizing all transactions correctly. If fraud systems are trained to identify fraud or non-fraud transactions, then the ensemble may specialize weights more effectively and holistically improve. 

\FloatBarrier
\begin{figure}[!ht]
\label{fig:TunedPerf}
\centering 
\includegraphics[width=1\linewidth]{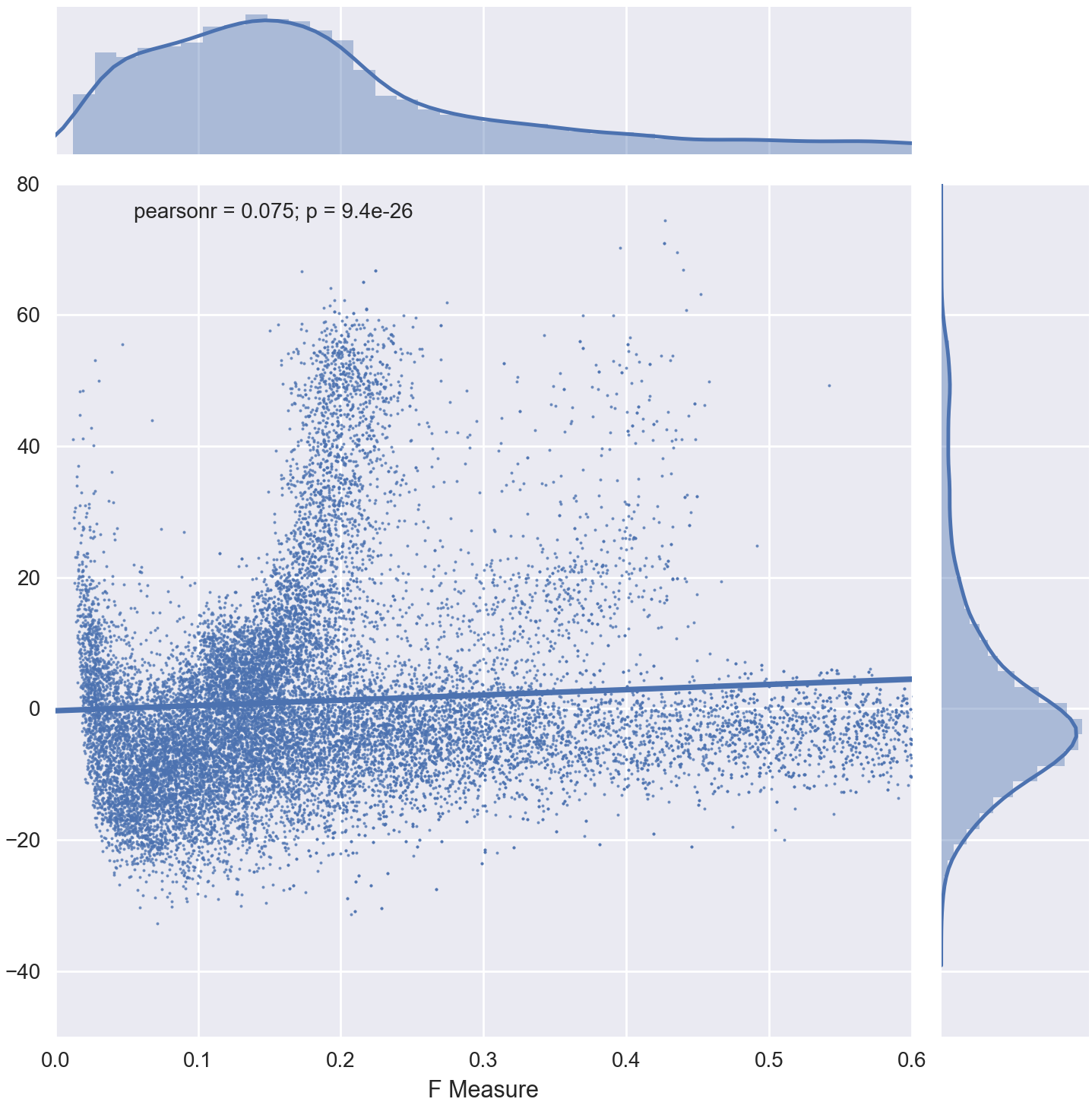}
\caption{Scatter plot of $F_1$ Score against fraud cost to identify relationship between performance indicators. The $y-axis$ represents fraud cost in USD - thousands.}

\end{figure}
\FloatBarrier

\subsection{$F_1$ Score uncorrelated with fraud cost performance}

%Outline the main reason for checking
While examining ensemble performance, it became clear that the $F_1$ Score did not depict any clear relationship with cost performance. However, many researchers including \cite{DalPozzolo2015} and \cite{Yen2009} cite the $F_1$ Score as a commonly used evaluation metric with imbalanced data problems. To garner a better perspective of the relationship between the $F_1$ Score to other metrics in this study, a scatter plot mapping it to fraud cost for all algorithms is shown in Figure 10. Since a higher $F_1$ Score indicates better predictive power, we expected to find a negative relationship between $F_1$ and fraud cost. 

%What we actually found for the F-measure
Instead, a slight positive relationship was found between $F_1$ and fraud cost. Little trending could be found using a linear model, and the $r$ value of the function mapping the plot is 0.071. At the same time, Figure 8 outlines how a single $F_1$ score often corresponds a wide range of costs. This is likely explained by the disproportionate weighting of $TPR$ over PPV. True positive rate considers how often fraud transactions are classified correctly and is calculated with $TP$ and $FN$, the two most heavily weighted detection outputs. Since $F_1$ combines both $TPR$ and PPV, a low $TPR$ and high $PPV$ can generate the same $F_1$ as a high $TPR$ and a low $PPV$. 

%Counter point about why to include PPV at all
Why, then, include $PPV$ in FDS evaluation at all? Without $PPV$, a model ceases to consider the occurrence of false positives (false alarms). In this study, $FP$ was weighted extremely low, as it only represents the cost of addressing a false alarm. However, it may cost a great deal more through its intangible effects on business operations. For example, a credit card user who consistently has their transactions blocked unnecessarily will likely stop using their card, losing companies revenue and perhaps tarnishing its public image. From another perspective, employees will be overwhelmed by false alarm transactions  until potentially fraudulent transactions can be verified automatically. Thus, $F_1$ necessarily considers $TPR$ and $PPV$, but appears unable to appropriately weight their influence for credit card fraud detection.

%What should my last paragraph be?
Another valid question is how to proceed with evaluating Fraud Detection Systems. It would be a mistake to propose quantifying performance solely with cost-based algorithms. The financial effects of fraud, particularly credit card fraud, are difficult to quantify in general, let alone for specific firms. An approximation-based cost matrix may add value to discerning optimal FDS practices, but ultimately it should be supplemented by other metrics. As with ensemble predictions, evaluation metrics are most effective when used in conjunction.

\section{Discussion}

%Add a few more sources about the topic at hand
The objectives of this analysis were to determine optimal approaches for addressing credit card fraud in online channels. As technology evolves, innovate methods of payment will continue to arise, potentially using blockchains or other systems. Therefore, it is imperative to develop a structure for detecting fraudulent purchases where little or no precedent exists and traditional approaches like profiling purchasing behavior are unavailable. 

% Sampling discussion
For some like \cite{Chawla2002}, the issue presents itself as a matter of sampling. Overcoming data imbalance is one of the major hurdles for efficiently preventing fraud. Therefore, we explicitly analyzed the benefits and drawbacks of undersampling and SMOTE. While it appeared that SMOTE's creation of synthetic records ultimately caused over-fitting to the minority class, researchers continue to add innovative adaptations \cite{han2005borderline}. Particularly with credit card fraud, the composition of training data is far more essential for success than size.

%Mode centric approach to detecting fraud
For others, credit card fraud detection is approached by finding and tuning the most effective model. While dozens of Fraud Detection Systems have been proposed and tested, few operate under the resource constraints found in industry \cite{Phua2010}. Therefore, we implemented five machine learning algorithms that are not reliant on computationally expensive procedures. The control model, K-Nearest Neighbors, embodies all of the pitfalls a traditional machine learning model can suffer from when applied to credit card fraud. Data imbalance tarnished KNN's predictive accuracy, even with re-balanced samples.

%Talk about the good models.
Fortunately, three models consistently generated low cost metrics. All models were scored on a cost matrix that emphasized the societal cost of fraud by transaction as opposed to generalizing a static cost for each fraud occurrence \cite{gadi2008credit}. Logistic regression proved to be the most efficient FDS and Random Forest was the most consistent across different samples. Implemented in another e-commerce fraud detection experiment, Logistic Regression's performance in \cite{maranzato2010fraud} found similar predictive success to its results in this study. Moreover, SVC classification was the second most effect FDS, despite its lack of a kernel function. Notably, it was able to do so in a fraction of the time necessary to run a full Support Vector Machine.

%Discuss ensemble techniques
Moreover, ensemble techniques and genetic algorithms are applied to fraud detection, because of their versatility in broad solution spaces and general ability to converge on an optimal solution. Duman \& Ozcelik (2011) \cite{DUMAN2011} found relative success detecting bank fraud with an ensemble of scatter search heuristics and genetic algorithm optimization. However, they score the performance of these algorithms solely on statistical metrics like AUROC, hit rate, and gini index. While these metrics are particularly useful for cross-study comparisons, they provide little information to companies that wish to minimize cost.

%Discuss the importance of cost
In 1997, Stolfo et. al. \cite{stolfo1997credit} proclaimed that  ``a cost function that takes into account both the True and False Positive rates should be used to compare ... classifiers." Many studies echo this sentiment, but some continue to publish innovative Fraud Detection Systems without contextualizing cost benefits. While the difficulty of quantifying the cost of fraud is a noted and worthy criticism, it does not reduce the potential benefits an approximated cost structure can provide as a supplemental metric for existing methods. $F_1$ scores were found to be uncorrelated with cost-based performance in this study. While it is likely no one would score a fraud detection algorithm on $F_1$ alone, the problems the metric suffers from are endemic to other measures. 

%Discuss potential problems with AUROC and AUPRC
Criticism of AUROC is often borne from its inclusion of $TN$ rates that may hold $FPR$ artificially low. Conversely, AUPRC metrics are commended for eliminating some of the effects of data imbalance by not considering $TN$. Yet, it is intuitively clear that unnoticed fraud transactions ($FN$) are far more detrimental to a company than false alarms ($FP$). Yet, AUPRC, a combination of these metrics, weights $TPR$ and $PPV$ equally. Shouldn't the area gained by improving $TPR$ hold a greater significance than area gained by $PPV$? We maintain it should.

%Back off a little to discuss how fraud cost should be used
With that said, the difficulty in discerning fraud cost mandates that it be supplemented by traditional metrics for several reasons. Most importantly, standardized metrics unrelated to cost allow for an unbiased categorization of Fraud Detection Systems from an academic standpoint. Furthermore, just as purchasing behavior changes over time, so do the costs of fraud to a firm. Fraud cost matrices for companies have a shelf life, and should only be used to supplement independent statistical measurements. In this way, an algorithm can be characterized independently and benchmarked against its peers, as well as provide contextual information to firms that may wish to implement it. The problem of transferring Fraud Detection Systems from academia to industry is continues today.

%Conclusion paragraph
Few implementation practices are considered empirically standard in the field of credit card fraud detection. Innovative Fraud Detection Systems continue to be discovered at an accelerating rate. To ensure these innovations positively impact the public, its performance must be clear to all parties involved. No matter how creative, detection algorithms that rely on only cost-based or statistical metrics \cite{Stolfo2000} often do not provide enough perspective to companies or researchers to quantify their impact. Evaluating fraud models in a holistic manner facilitates the adoption of Fraud Detection Systems worldwide, increasing the probability that preventative technology outpaces the proliferation of credit card fraud as spreads to new channels.

%%%%%%%%%%%%%%%%%%%%%%%%%%%%%%%%%%%%%%%%%%%%%%%%%%%%%%%%%%%%%%%%%%%%%%%
%	End OF THE APPENDIX

%% References

% % %	Bibliography
% % %
% % %%%%%%%%%%%%%%%%%%%%%%%%%%%%%%%%%%%%%%%%%%%%%%%%%%%%%%%%%%%%%%%%%%%%%%%
\clearpage
\pagebreak
\newpage
\bibliographystyle{model1-num-names}
\bibliography{FraudCitations.bib}

% % %%%%%%%%%%%%%%%%%%%%%%%%%%%%%%%%%%%%%%%%%%%%%%%%%%%%%%%%%%%%%%%%%%%%%%%

\newpage
\appendix

\end{document}